\documentclass[10pt]{article} 
\usepackage[accepted]{tmlr}

\usepackage[utf8]{inputenc} 
\usepackage[T1]{fontenc}    
\usepackage{url}            
\usepackage{booktabs}       
\usepackage{amsfonts}       
\usepackage{nicefrac}       
\usepackage{microtype}      
\usepackage{xcolor}         

\usepackage{graphicx}
\usepackage{wrapfig}
\usepackage{booktabs} 
\usepackage{enumitem}
\usepackage{bbding}
\usepackage[font=small]{caption}
\usepackage{multirow}
\usepackage{listings}
\usepackage{url}
\usepackage{amsmath}
\usepackage{amssymb}
\usepackage{mathtools}
\usepackage{amsthm}
\usepackage{algorithm}
\usepackage{algpseudocode}
\usepackage{bm}
\usepackage{multirow}
\usepackage{lscape}
\usepackage{longtable}
\usepackage{tabularx}

\definecolor{cb_orange}{RGB}{213,94,0}
\definecolor{cb_green}{RGB}{34,136,51}
\definecolor{cbgreen}{RGB}{34,136,51}
\definecolor{sky_blue}{RGB}{204, 238, 255}
\definecolor{cb_purple}{RGB}{170, 51, 119}
\definecolor{cb_red}{RGB}{204, 51, 17}
\definecolor{cb_blue}{RGB}{0, 119, 187}
\definecolor{cbblue}{RGB}{0, 119, 187}
\definecolor{mydarkblue}{rgb}{0,0.08,0.45}

\definecolor{codegreen}{rgb}{0,0.6,0}
\definecolor{codegray}{rgb}{0.5,0.5,0.5}
\definecolor{codepurple}{rgb}{0.58,0,0.82}
\definecolor{backcolour}{rgb}{0.95,0.95,0.92}
\definecolor{lightgray}{rgb}{0.8,0.8,0.8}

\lstdefinestyle{mystyle}{
    backgroundcolor=\color{backcolour},   
    commentstyle=\color{codegreen},
    keywordstyle=\color{magenta},
    numberstyle=\tiny\color{codegray},
    stringstyle=\color{codepurple},
    basicstyle=\ttfamily\footnotesize,
    breakatwhitespace=false,         
    breaklines=true,                 
    captionpos=b,                    
    keepspaces=true,                 
    numbers=left,                    
    numbersep=5pt,                  
    showspaces=false,                
    showstringspaces=false,
    showtabs=false,                  
    tabsize=2
}
\usepackage[colorlinks=true,citecolor=cbgreen,linkcolor=cbblue,urlcolor=cbblue]{hyperref}
\usepackage[capitalize,noabbrev]{cleveref}

\newcommand{\update}[1]{{\color{black}{#1}}}

\newcommand{\compwob}{CompWoB}

\title{Exposing Limitations of Language Model Agents in Sequential-Task Compositions on the Web}

\author{
    \name Hiroki Furuta$^{1,2}$\thanks{Work done as Student Researcher at Google DeepMind.} \quad Yutaka Matsuo$^{2}$ \quad Aleksandra Faust$^{1}$ \quad Izzeddin Gur$^{1}$ \\
    \email \{furuta,matsuo\}@weblab.t.u-tokyo.ac.jp \quad \{sandrafaust,izzeddin\}@google.com \\
    \addr $^{1}$Google DeepMind \quad $^{2}$The University of Tokyo
}

\def\month{12}  
\def\year{2024} 
\def\openreview{\url{https://openreview.net/forum?id=Y9kAsYIjYc}} 

\begin{document}

\maketitle

\begin{abstract}
Language model agents (LMA) recently emerged as a promising paradigm on muti-step decision making tasks, often outperforming humans and other reinforcement learning agents.
Despite the promise, their performance on real-world applications that often involve combinations of tasks is still underexplored.
In this work, we introduce a new benchmark, called \textit{\compwob{}} -- 50 new compositional web automation tasks reflecting more realistic assumptions.
We show that while existing prompted LMAs (\texttt{gpt-3.5-turbo} or \texttt{gpt-4}) achieve 94.0\% average success rate on base tasks, their performance degrades to 24.9\% success rate on compositional tasks.
On the other hand, transferred LMAs (finetuned only on base tasks) show less generalization gap, dropping from 85.4\% to 54.8\%.
By balancing data distribution across tasks, we train a new model, HTML-T5++, that surpasses human-level performance (95.2\%) on MiniWoB, and achieves the best zero-shot performance on \compwob{} (61.5\%).
While these highlight the promise of small-scale finetuned and transferred models for task compositionality, their performance further degrades under different instruction compositions changing combinational order.
In contrast to the recent remarkable success of LMA, our benchmark and detailed analysis emphasize the necessity of building LMAs that are robust and generalizable to task compositionality for real-world deployment.
\end{abstract}

\section{Introduction}
Based on the exceptional capability of large language models (LLMs)~\citep{openai2023gpt4,anil2023palm2,touvron2023llama} in commonsense understanding~\citep{brown2020gpt3,Chowdhery2022palm}, multi-step reasoning~\citep{wei2022cot,kojima2022lets}, program synthesis~\citep{chen2021evaluating} and self-improvement~\citep{shinn2023reflexion,madaan2023selfrefine,to2023better}, language model agents (LMAs) have recently emerged to tackle various decision making problems, such as robotics~\citep{huang2022language,ahn2022saycan}, information retrieval~\citep{nakano2021webgpt,yao2022react}, and external tool use~\citep{wu2023visual,shen2023hugginggpt}.
Especially, \textit{web automation}~\citep{shi2017miniwob} has attracted attention a lot as an promising application of LMAs, because LMAs with prompting~\citep{kim2023rci,sun2023adaplanner,zheng2023synapse} outperform humans and other learning-based agents, such as imitation~\citep{furuta2023mmwebnav} or reinforcement learning~\citep{humphreys2022data}.

Despite their human-level proficiency in MiniWoB~\citep{shi2017miniwob}, a standard web automation benchmark, it is still unclear whether LMAs could deal with challenges in the real world; such as complex observation~\citep{gur2023realworld}, domain generalization~\citep{deng2023mind2web}, and ambiguity of instructions~\citep{zhou2023webarena}.
These challenges are exacerbated due to the open-ended nature of real-world tasks, making it infeasible for the agent system to prepare exemplars and prompts for any unseen task in advance.
Moreover, it is reasonable and practically important to assume the \textbf{zero-shot evaluation of the off-the-shelf agents at deployment}. This is an orthogonal assumption to the recent progress in LMAs learnable from execution/self-feedback~\citep{shinn2023reflexion,ma2023laser} during the deployment. Considering the agents are deployed as a service, (1) it is infeasible to cover all the possible prompts/demonstrations for user requests in advance and (2) iterative trial-and-error to adjust system prompts hurts user experiences.

\begin{figure*}[t]
\begin{minipage}[c]{0.63\textwidth}
    \centering
    \includegraphics[width=\linewidth]{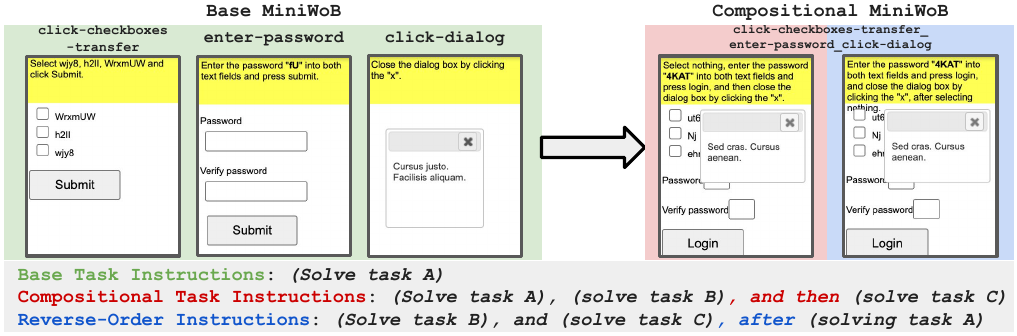}
\end{minipage}
\begin{minipage}[c]{0.35\textwidth}
    \centering
    \includegraphics[width=\linewidth]{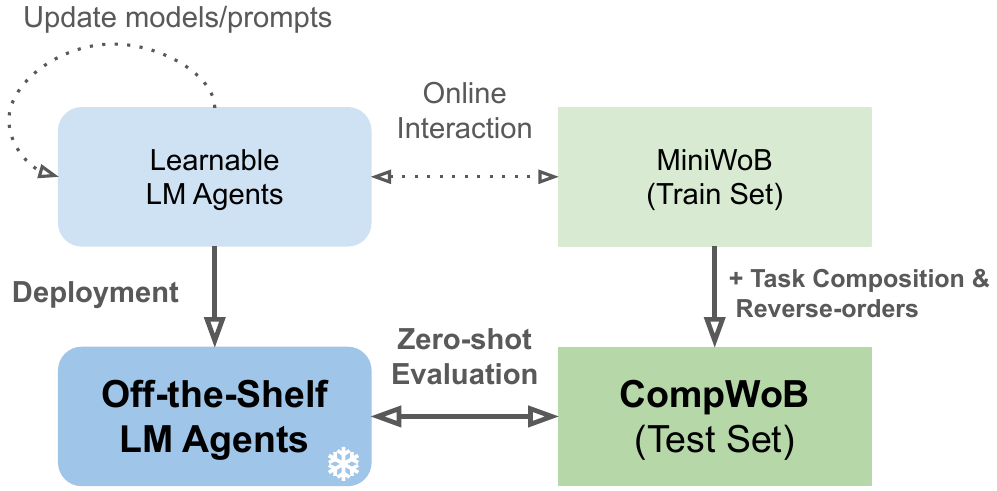}
\end{minipage}
\vskip -0.1in
\caption{
(\textbf{Left})
We design \textit{\compwob{}}, as a novel decision making benchmark for LMAs, by leveraging the high composability of simulated web environments.
We first select base tasks from the original MiniWoB~\citep{shi2017miniwob} based on the brute-force task complexity averaged among existing LMAs. 
While considering the feasibility and reality, we systematically combine them into a sequential single task (e.g. \texttt{click-checkboxes-transfer} + \texttt{enter-password} + \texttt{click-dialog}$\rightarrow$ \texttt{click-checkboxes-transfer\_enter-password\_click-dialog}). The instructions of base tasks are stitched with ``\textbf{and then}''. LMAs are asked to satisfy the given instructions sequentially (e.g. satisfying the success criteria of task \textit{A} $\rightarrow$ \textit{B} $\rightarrow$ \textit{C}).
We also implement \textbf{reverse-order instruction} settings, where the instructions are provided upside down (e.g. solve task \textit{B}, and solve task \textit{C}, after solving task \textit{A}). These complex yet controllable strategies make the analysis of LMA's behaviors easy, while maintaining complex and ambiguous aspects of real-world web environments.
(\textbf{Right}) In this paper, we assume the zero-shot evaluation of the off-the-shelf agents at deployment, which is practically important because (1) it is infeasible to cover all the possible prompts/demonstrations for user requests in advance and (2) iterative trial-and-error to adjust system prompts hurts user experiences.
\compwob{} works as a hold-out test environment accompanying with MiniWoB as a train environment.
}
\vskip -0.2in
\label{fig:compositional_miniwob}
\end{figure*}

In this work, we extensively study the transferability of LMAs to more realistic sequential task compositions.
We first design a new controlled test bed, called \compwob{}, with 50 compositional tasks by combining a set of base tasks based on their difficulty~(\autoref{fig:compositional_miniwob}, right).
\compwob{} works as a \textit{hold-out} test environment accompanying with MiniWoB as a train environment~(\autoref{fig:compositional_miniwob}, left).
Each compositional task is implemented from 2 to 8 base tasks in a single-page or multi-page environment with instructions linked together using simple connectors such as ``and then''.
Only providing the knowledge about base tasks, we investigate the generalization performance of existing SoTA prompted LMAs~\citep{kim2023rci,sun2023adaplanner,zheng2023synapse} with planning, self-improvement, program synthesis, and structured prompts that are supported by \texttt{gpt-3.5-turbo} and \texttt{gpt-4}.
Our findings indicate that their performance drops significantly, from 94.0\% success on base tasks to 24.9\% success on compositional tasks.
In contrast, small-scale LMAs finetuned only on base tasks and zero-shot-transferred to compositional settings (i.e. transferred LMAs), deal with unknown task compositionality better, achieving 54.8\% success rate on average.
By rebalancing the data distribution, we train a new model, HTML-T5++, that achieves human-level performance on MiniWoB and performs the best among all the LMAs on compositional tasks.
In contrast, small-scale LMAs finetuned only on base tasks and zero-shot-transferred to compositional settings (i.e. transferred LMAs), deal with unknown task compositionality better, achieving 54.8\% success rate on average.
By rebalancing the data distribution, we also train a new model, HTML-T5++, that achieves human-level performance on MiniWoB and performs the best among all the LMAs on compositional tasks.
We further point out that LMAs struggle to handle complex instruction compositions permuting the order of sub-instructions, where prompted agents are more robust to the difference in the order of compositions compared to transferred agents (6.9\% vs 23.8\% drop in performance).
Finally, we illustrate that instruction length and observation complexity are useful indicators of compositional task performance.

In contrast to the recent notable success of LMAs, our benchmark and detailed analysis highlight building robust and generalizable LMAs to be safely deployed in the real world.
In summary, our key contributions are:
\begin{itemize}[leftmargin=0.5cm,topsep=0pt,itemsep=0pt]
    \item We empirically show that (1) \textbf{prompted LMAs even with} \texttt{gpt-4} \textbf{suffer from generalizing to compositional web automation tasks much more than transferred LMAs}, and (2) \textbf{LMAs are highly sensitive to the order of instructions}.
    \item We develop \textit{\compwob{}}
    \footnote{\url{https://github.com/google-research/google-research/tree/master/compositional_rl}}
    , simulated web environments for LMAs to measure the generalization to the realistic task compositionality and complex instructions.
    \item We explore a new data mixture heuristic for finetuning LMAs, where we find that it is effective to add demonstrations of the tasks that suffer from data shortage and then gradually reduce the ratio of easier tasks to focus more on challenging tasks. HTML-T5++, trained on our novel heuristic, achieves human-level performance on MiniWoB (95.2\%) and the best zero-shot transfer to \compwob{} (61.5\%).
    \item Our experiments reveal that while more capable models, such as \texttt{gpt-4}, lead to better performance, they still struggle with task compositionality or complex instructions.
\end{itemize}

\section{Related Works}
\textbf{Web Automation}~
Although prior works have worked on imitation learning and reinforcement learning~\citep{liu2018wge,gur2018learning,jia2018domqnet,humphreys2022data}, web automation has become a popular domain as an application of LMAs~\citep{gur2022html,kim2023rci}.
In earlier work, finetuned LMAs, based on at most 3-billion parameters, amortize the training costs with the strong prior knowledge on web environments~\citep{gur2022html,furuta2023mmwebnav,shaw2023pixels}, but they often result in sub-optimal performances due to the insufficient data coverage.
Recently, by leveraging capable private LLMs~\citep{brown2020gpt3,ouyang2022instructgpt} with self-refinement~\citep{kim2023rci}, program synthesis~\citep{sun2023adaplanner}, structured instruction-state translation~\citep{zheng2023synapse}, or hierarchical prompts~\citep{sridhar2023hierarchical,ma2023laser}, prompted LMAs with few-shot exemplars have outperformed finetuned LMAs and shown superior performance to humans and RL-finetuned agents.
In contrast, our work discusses the transferability and robustness of those LMAs in zero-shot web automation with a set of compositional tasks, and resolves the sub-optimality of finetuned LMAs via data-rebalancing.

In addition to MiniWoB~\citep{shi2017miniwob}, a representative web simulator, several works have conducted real-world evaluation~\citep{gur2023realworld} and proposed novel benchmarks reflecting real-world assumptions, such as a simulated e-commerce site~\citep{yao2022webshop}, sand-boxed real-world websites~\citep{zhou2023webarena,workarena2024}, an adaptive sim-to-real bridge with unsupervised auto-curricula~\citep{gur2021gminiwob}, and large-scale web interaction dataset curated by human annotators~\citep{deng2023mind2web}.
However, real-world web automation may make the analysis challenging because it often faces miscellaneous obstacles, such as complex HTML observations, domain gaps between websites, and ambiguous instructions.
In this work, we design \compwob{} under realistic assumptions while controlling
task difficulty and ambiguity of instructions, and investigate what may prevent the generalization capability of LMAs in compositional tasks.


\textbf{Language Model Agents}~~
Beyond the common NLP tasks, LLMs could act as autonomous agents~\citep{wang2023survey,qin2023tool} to solve the given instruction-following tasks, by considering the context in the prompt as states~\citep{ahn2022saycan,yao2022react,hsieh2023tool} and sequentially planning and manipulating external ``tools'' or ``actuators'', such as calculators~\citep{parisi2022talm}, retrievers~\citep{schick2023toolformer,hao2023toolkengpt}, APIs~\citep{qin2023toolllm,tang2023toolalpaca}, programs~\citep{gao2023pal,wang2023voyager,liang2023code,song2023restgpt,cai2023large}, robotic commands~\citep{huang2022language,huang2022inner,tang2023saytap}, computer game~\citep{nottingham2023embodied,wang2023describe}, or other foundation models~\citep{lu2023chameleon,hsieh2023tool,wu2023visual,shen2023hugginggpt,yang2023gpt4tools}. Those prior works have worked on proposing novel benchmarks~\citep{li2023apibank,xu2023tool,patil2023gorilla} and comparing backbone LLMs (e.g. open-sourced v.s. private)~\citep{ruan2023tptu,liu2023bolaa,liu2023agentbench}.
Despite their success, it is still unclear how such LMAs designed for specific tasks can generalize out-of-domain problems, which should be an important perspective since we may not prepare prompts and exemplars for all the possible problems in the real world.
In this work, we measure the transferability and robustness to unseen compositionality and instructions in a realistic web automation scenario.

\textbf{Large Language Models for Compositional Tasks}~
Several works have investigated compositional natural language problems with LLMs, such as semantic parsing~\citep{furrer2021compositional,shaw2021compositional,zhou2023leasttomost}, logic grid puzzles~\citep{dziri2023faith}, mathematical reasoning~\citep{chen2023skillsincontext}, programming~\citep{zelikman2023parsel}, and planning~\citep{brahman2023plasma}, which shows that dynamical selection of exemplars for decomposed sub-problems~\citep{drozdov2023compositional} or model scaling~\citep{qiu2022evaluating} could help generalization.
While those are focused on static tasks, our paper deals with task compositionaliy in interactive decision making, especially, in web automation where the task may have more explicitly decomposable structures~\citep{gur2021gminiwob} than natural language tasks.

\section{Preliminaries}
Web automation could be described as a deterministic sequential decision making problem, which consists of a state space~$\mathcal{S}$, action space~$\mathcal{A}$, deterministic transition function $T: \mathcal{S} \times \mathcal{A} \xrightarrow{} \mathcal{S}$, a set of instructions $\mathcal{G}$, a set of contextual information (i.e. prompts for LLM) $\mathcal{C}$, and episodic reward function (i.e. success criteria)~$r: \mathcal{S} \times \mathcal{G} \times \mathcal{A} \xrightarrow{} \{0, 1\}$.
At each time step $t$, the language model agent $\pi$ infers the action conditioned on the prompt, instruction, current state, and previous actions $\pi:  \mathcal{S} \times \mathcal{A}^{\times t} \times \mathcal{C} \times \mathcal{G} \rightarrow{} \mathcal{A}$, and moves to the next state: $s_{t+1} = T(s_t, a_t)$.
When the agent reaches the terminal state (e.g. \texttt{Login} button is clicked) or the max time step is exceeded, the episode is marked as a success if the instruction $g$ is satisfied (i.e. $r(s_t, g, a_t) = 1$).
The state $s_t \in \mathcal{S}$ is a raw HTML, and we assume the programmatic action space: \texttt{function(selector, text)}.
\texttt{function} is either \textit{click}, \textit{move} or \textit{type}, \texttt{selector} is an integer index or XPath that can uniquely specify the element, and \texttt{text} is a text input for \textit{type} function.

\textbf{Task Compositionality}~
Web automation tasks can be decomposed into a set of primitive base tasks. For instance, (1) clicking several checkboxes, (2) fulfilling the password form, and (3) closing the dialog window. Such a combination could be open-ended and have some dependencies. In this work, we assume that the task $\psi \in \Psi$ is characterized by a corresponding subtree of HTML ($\mathcal{S}_{\psi} \subset \mathcal{S}$) and instructions ($\mathcal{G}_{\psi} \subset \mathcal{G}$), and can be dependently combined each other as long as the compositional task is executable.

\section{Language Model Agents for Web Automation}

To be self-contained, we here review the representative LMAs, which are selected based on their superior performance and novelty in using LLMs for web automation; RCI (Section~\ref{sec:lma_rci}), AdaPlanner (Section~\ref{sec:lma_adaplanner}), Synapse (Section~\ref{sec:lma_synapse}).
These are characterized with different base LLMs, prompting and pipeline.
To clarify their methodological difference, we provide the pseudo code in \autoref{sec:lma_pseudo_code}.
Moreover, we resolve the sub-optimal performance of transferred LMAs (Section~\ref{sec:finetuned_lma}) with data-rebalanced finetuning (Section~\ref{sec:html_t5_++}).

\subsection{RCI}
\label{sec:lma_rci}

The agent with Recursive Criticism and Improvement (RCI) prompting~\citep{kim2023rci} first generates an open-loop plan to follow a given instruction using few-shot demonstrations. Next, it uses a prompt-based critic to identify the errors in the plan and improves it by reflecting on self-criticized outputs, which is referred as an explicit RCI (ERCI) loop.
After ERCI, the agent follows the self-improved plan step-by-step. Before executing the action at each step, the agent grounds the action to the current state (i.e. HTML, open-loop plan, and previous actions) and refines its formatting to be parsable, which increases the feasibility and reduces hallucinations.
These final steps are referred as an implicit RCI (IRCI) loop without the self-criticism.
All of those play complementary roles to achieve human-level proficiency.
While 1 ERCI and 3 IRCI loops are recommended, we observe that the optimal number of self-improvement iterations may differ across the tasks (see Appendix~\ref{sec:hyperparams_rci}).

\subsection{AdaPlanner}
\label{sec:lma_adaplanner}
In contrast to other LMAs, AdaPlanner~\citep{sun2023adaplanner} leverages the capability of program synthesis in LLMs to mitigate the hallucination in a plan. Conditioning on the instruction, the description of permissible actions in the web environments, and few-shot demonstrations, the agent first generates an open-loop plan in a Python function, where each snippet corresponds to the action. Once the agent receives environmental feedback at each step, such as assertion errors in the code, other functional errors, or ``ask'' action to LLMs, it adaptively re-generates the plan for the remaining steps in a closed-loop manner. LLMs more capable of code have performed better, such as \texttt{text-davinci-003} than \texttt{gpt-3.5-turbo}.

\begin{wraptable}{R}[0pt]{0.4\linewidth}
\begin{center}
\begin{small}
\scalebox{0.675}{
\begin{tabular}{llr}
\toprule
\textbf{Models} & \textbf{Dataset Size} & \textbf{Success Rate} \\
\midrule
HTML-T5 & 347K episodes & 85.6\% \\
\midrule
 & 424K (+77K)  & 94.1\% \\
\textbf{HTML-T5++} & 351K (+77K - 73K)  & 94.6\% \\
\multicolumn{1}{c}{(ours)} & 322K (+77K - 102K)  & 94.8\% \\
 & \textbf{282K (+77K - 142K)}  & \textbf{95.2}\% \\
 & 241K (+77K - 183K)  & 95.0\% \\
\midrule
\midrule
\multicolumn{2}{l}{RCI~\citep{kim2023rci}} & 90.6\% \\
\multicolumn{2}{l}{AdaPlanner~\citep{sun2023adaplanner}} & 92.9\% \\
\multicolumn{2}{l}{Human} & 93.5\% \\
\multicolumn{2}{l}{CC-Net~\citep{humphreys2022data}} & 93.5\% \\
\multicolumn{2}{l}{RCI (\texttt{gpt-4}) ~\citep{kim2023rci}} & 94.0\% \\
\multicolumn{2}{l}{Synapse~\citep{zheng2023synapse}} & \textbf{98.5}\% \\
\bottomrule
\end{tabular}
}
\end{small}
\end{center}
\vskip -0.15in
\caption{
Average success rate of finetuned LMAs in 56 tasks on MiniWoB.
Adding 77K episodes and reducing redundant thousands of episodes, HTML-T5++ achieves competitive performance to prompted LMAs, RL-finetuned agents, and humans, while improving the success rate from 85.6\% to 95.2\%.
}
\vskip -0.25in
\label{tab:data_rebalancing}
\end{wraptable}

\subsection{Synapse}
\label{sec:lma_synapse}
Synapse~\citep{zheng2023synapse} argues that LMAs perform better if well-designed structured prompts are provided, even without self-improvement or program synthesis.
The structured prompting is formed by two pre-processing strategies: state filtering and task reformulation. State filtering gradually transforms raw HTML into simple formatted text, such as a Pythonic list or dictionary, in a multi-step manner, which may improve the state understanding of LMAs. Task reformulation translates given instructions or raw HTML into decomposed queries: for instance, translating \textit{``select 12/03/2016 as the date and hit submit''} into \textit{``select the datepicker at step 1, click 'Prev' 7 times at step 2-8 (May is 7 months before December), click the date '12' at step 9, and finally submit at step 10''} (translated instruction), or mapping proper noun into corresponding XPath (translated HTML).
While detailed structured prompts have led to strong performances, those should be specialized for each primitive task in MiniWoB by leveraging 7 different types of reformulation.
See Appendix~\ref{sec:hyperparams_synapse} for further details.

\subsection{Finetuned and Transferred Language Model Agents}
\label{sec:finetuned_lma}

In addition to the prompted LMAs, LMAs finetuned on base tasks have also been developed~\citep{gur2022html,furuta2023mmwebnav,shaw2023pixels}, which are built on pre-trained language models, such as T5~\citep{2020t5}, Flan-T5~\citep{chung2022flant5}, HTML-T5~\citep{gur2023realworld}, or Pix2Struct~\citep{lee2023pix2struct}, with web automation demonstrations.
Those LMAs take HTML (or screenshots) and previous actions as inputs and predict the text-format next actions in a closed-loop manner.
Since pre-trained language models have a sufficient inductive bias for web environments and instruction-following, finetuned LMAs can data-efficiently achieve competitive performance to the RL-finetuned agents trained from scratch with domain-specific architectures~\citep{humphreys2022data,liu2018wge}.
Compared to the prompted LMAs relying on private LLM API, it is possible to build on-premise agents based on tractable-size models (at most 3 billion parameters), which may reduce inference time and costs.
However, prior works have pointed out that finetuned LMAs struggle to the sub-optimal performance~\citep{zheng2023synapse} and they require demonstrations on the order of hundreds of thousands while prompted LMAs just need hundreds of episodes~\citep{kim2023rci}.
In this paper, we extensively evaluate such finetuned LMAs in \textbf{zero-shot transfer} settings; LMAs are finetuned only with base task demonstrations and should deal with unseen compositional tasks.
We call those \textit{transferred} LMAs in the later sections.

\subsection{Data-Rebalancing Improves Finetuned Language Model Agents}
\label{sec:html_t5_++}

\citet{furuta2023mmwebnav} utilized agent-driven data collection instead of humans to improve the performance of finetuned LMAs further.
For each task, 10k demonstrations are collected and filtered based on task success, which resulted in challenging tasks having much less than 10k demonstrations due to the sub-optimal performance of LMA on these tasks.
We identify that by fixing the data-imbalance problem, the performance of finetuned LMAs can be significantly improved, achieving super-human performance on MiniWoB.
We first run Synapse~\citep{zheng2023synapse} on MiniWoB and collect 77K additional demonstrations across 16 tasks on top of 347K demonstrations~\citep{furuta2023mmwebnav} to compensate for the lack of data in specific tasks.
We then estimate the ``brute-force'' task difficulty averaging success rates for representative web automation agents. Based on those proximal measures, we classify 65 base tasks into three categories, such as \texttt{easy} (0.8 - 1.0), \texttt{medium} (0.6 - 0.8), and hard (0.0 - 0.6) (see \autoref{sec:task_complexity}).
We then balance the number of episodes based on the task difficulty, where we gradually reduce the ratio of easier tasks to focus more on challenging tasks. For instance, we remove $X$\% episodes from top-$k$ tasks in \texttt{easy} group. We heuristically design the following data-mixing strategies:
\begin{itemize}[leftmargin=0.5cm,topsep=0pt,itemsep=0pt]
    \item Original dataset released by~\citet{furuta2023mmwebnav} (347K episodes).
    \item Adding 77K episodes from 16 tasks suffering the lack of data~(424K episodes).
    \item Removing 50\% episodes from top-10 \texttt{easy} tasks (424K - 73K = 351K episodes)
    \item Removing 80\% episodes from top-10 \texttt{easy} tasks (424K - 102K = 322K episodes)
    \item Removing 50\% episodes from \texttt{easy} tasks (424K -142K = 282K episodes)
    \item Removing 80\% episodes from top-15 \texttt{easy} tasks and removing 50\% episodes from other 11 \texttt{easy} tasks (424K - 183K = 241K episodes)
\end{itemize}
See \autoref{sec:data_rebalance_details} for further details.
The automation of data mixture design~\citep{xie2023doremi} would be an important future direction.

We finetune HTML-T5-XL~\citep{gur2023realworld}, a pre-trained language model with local and global attention in the encoder and a mixture of long-span denoising, on these rebalanced datasets.
\autoref{tab:data_rebalancing} shows that all the data-rebalance strategies improve the success rate, and reducing 50\% episodes from \texttt{easy} tasks
(finally 282K episodes in total) is the most effective rebalancing strategy.
This suggests that finetuned LMAs can be as capable as prompted LMAs in decision making tasks.
\update{Moreover, due to the inherent compositional nature in challenging web automation tasks, data rebalancing may improve the capability of dealing with unseen task compositions.}
We include HTML-T5++ as a baseline in the later sections.

\begin{table}[tb]
\begin{center}
\begin{small}
\scalebox{0.925}{
\begin{tabular}{lllllllllll}
\toprule
 & \multicolumn{5}{c}{\textbf{\# of Episode Steps}} & \multicolumn{5}{c}{\textbf{\# of Instruction Tokens}} \\
\cmidrule(r){2-6} \cmidrule(r){7-11}
\textbf{Benchmark} & \textbf{Average} & \textbf{50th} & \textbf{90th} & \textbf{95th} & \textbf{Max} & \textbf{Average} & \textbf{50th} & \textbf{90th} & \textbf{95th} & \textbf{Max}\\
\midrule
MiniWoB~\citep{shi2017miniwob} & 3.6 & 4.3 & 5.5 & 5.8 & 7.0 & 8.5 & 14.8 & 30.5 & 12.9 & 23.2\\
Mind2Web~\citep{deng2023mind2web}& 7.7 & 6.0 & 14.0 & 17.0 & 37.0 & 19.2 & 16.0 & 33.0 & 42.0 & 84.0\\
CompWoB~(\textbf{Ours}) & 7.2 & 6.8 & 12.8 & 14.2 & 17.8 & 37.3 & 34.2 & 55.2 & 57.9 & 100.0\\
\bottomrule
\end{tabular}
}
\end{small}
\end{center}
\vskip -0.15in
\caption{
    The statistics (Mean, Max, 50/90/95th percentiles) of episode steps and instruction tokens from MiniWoB~\citep{shi2017miniwob}, Mind2Web~\citep{deng2023mind2web}, and CompWoB.
    CompWoB successfully increases task complexity from MiniWoB, and has sufficient difficulties the same as Mind2Web, a representative benchmark from real websites.
}
\label{tab:benchmark_comparison}
\end{table}

\section{Designing \compwob{} Controlling Task Complexity and Compositionality}

Even though MiniWoB includes a spectrum of simulated environments, they have still focused on narrow and single-task instances.
We need more advanced environments to measure the generalization to the various functional challenges in real-world web automation, such as complex observation~\citep{deng2023mind2web}, instruction~\citep{zhou2023webarena}, and task compositionality~\citep{gur2021gminiwob}.
We design \textit{\compwob{}} (i.e. compositional MiniWoB), as a general test bed for LMAs, by leveraging the high composability of simulated web environments (\autoref{fig:compositional_miniwob}).
In \compwob{}, we systematically combine several base tasks (from 2 to 8) in the original MiniWoB, which LMAs can already solve, into a single task (e.g. \texttt{click-checkboxes-transfer} + \texttt{enter-password} + \texttt{click-dialog} $\rightarrow$ \texttt{click-checkboxes-transfer\_enter-password\_click-dialog}).
This allows us to control the complexity of HTML and instructions, ensuring novel tasks are solvable to some extent.
Moreover, \compwob{} can work as a \textit{hold-out} test environment accompanying with MiniWoB as a train environment.
While tasks are visually simplified, CompWoB reflects a broader spectrum of functionalities in real-world websites without sacrificing controllability. For instance, the task in \cref{fig:compositional_miniwob} sketches the structure of a login form with agreement checkboxes and a pop-up window, like Internet banking. We also designed tasks with page transitions, such as from the login form to the email browser.

\subsection{Details of Task Design}
As mentioned in Section~\ref{sec:html_t5_++}, we first calculate the average success rates among representative web automation agents as brute-force task complexity, and classify 65 primitive tasks in MiniWoB into 3 categories (\texttt{easy}, \texttt{medium}, \texttt{hard}) based on those scores.
The details of classification are described in \autoref{sec:task_complexity}.
We randomly select base tasks from \texttt{easy} group, filter those combinations by their feasibility and reality, and make 50 compositional tasks. We divide those into five categories: \textbf{two-way} tasks (20), \textbf{three-way} tasks (10), \textbf{n-way} tasks (5), \textbf{transition} tasks (5), and \textbf{easy-medium two-way} tasks (10). In n-way tasks, we combine from 4 to 8 tasks sequentially, and in transition tasks, we implement explicit page transition, for instance, transiting from the login form to the email browser.
We also sample several tasks from \texttt{medium} group to construct easy-medium two-way tasks.
You can find the full list of tasks in \autoref{sec:per_task}.
For the implementation, we stitch the instructions with ``\textit{and then}'' and put each HTML on the same depth.
LMAs should satisfy the given instructions sequentially, such as from task \textit{A}, \textit{B}, to \textit{C}.
Moreover, to test whether LMAs can deal with complex and ambiguous instructions, we propose \textbf{reverse-order instruction} settings, where the instruction is provided upside down while its task order is semantically the same (e.g. solve task \textit{B} and \textit{C}, after solving \textit{A}).
These simple yet controllable strategies make the analysis of LMA's behaviors tractable while reflecting compositional aspects of real-world tasks.

\subsection{Number of Tasks}
MiniWoB has around 104 base tasks in total; all the two-way and three-way combinations of these tasks would give 185,460 compositional tasks. It is infeasible to manually curate all the two-way/three-way combinations. Unfortunately, it is also nontrivial to automate this process due to the locality of the environment implementations in MiniWoB. Each environment is mostly self-contained, making it nontrivial to modularize the whole benchmark/codebase so that every combination can be automatically generated. Alternatively, we outline a set of design principles -- solvability, feasibility, and reality, which we follow to manually curate 50 compositional tasks and 50 reverse-order instruction tasks that could cover a wide variety of difficulty, and compositionality. We believe these guiding principles would be applicable to other compositional generalization problems such as robotic navigation.
Based on these design principles, our compositional benchmark can easily be extended in the future to study even more challenging and compositional web automation problems.
We would also like to highlight that, because previous works were tested around 50 - 60 MiniWoB tasks~\citep{kim2023rci,sun2023adaplanner,zheng2023synapse,gur2022html,furuta2023mmwebnav}, 50 compositional tasks is a decent number of tasks to evaluate the agents. \update{Our addition of a hold-out compositional test set doubles the number of tasks for the community to study.}

\subsection{Inherent Compositionality and Sub-Task Dependency in Web Automation}
While many of the real-world tasks have inherent compositionality to some degree, these tasks are not explicitly designed for compositionality, making it difficult to systematically investigate the generalization gap. This can be analyzed with our CompWoB by separating the difficulty of the base tasks themselves and the difficulty of the task compositions.

Real-world websites have dependencies between sub-tasks. For instance, an email client requires a successful login to proceed to writing an email or reading social media posts of a specific user requires navigating to the personal page of that user. In CompWoB, we have implemented these kinds of dependencies in the reward function using logical AND operations. For instance, while we allowed agents to continue to ``write an email'' sub-task without successful login, the episode is flagged as a failure (i.e. zero reward) even if the ``write an email'' sub-task is successful. We inform agents of these dependencies using connectors in instructions such as ``solve B after A'' or ``solve A then B''. Our analysis in \cref{sec:task_complexity_main} also suggests that, in addition to task compositionality, long instructions and deep HTML sub-tree can lead to challenging tasks.

\subsection{Comparison to Other Web Automation Benchmarks}
We design CompWoB on top of MiniWoB, which enables us to test the human-level LMAs~\citep{kim2023rci,sun2023adaplanner,zheng2023synapse,gur2023realworld} as strong baselines. Because the success of in-domain tasks is ensured, we could focus on the analysis of generalization to unknown task compositions and their functionality in a controllable way. In contrast, other benchmarks copying real websites fail to obtain decent agents (for instance, the episode success rate of Mind2Web~\citep{deng2023mind2web} is around 5-10\% at most; around 15\% success at most in WebArena~\citep{zhou2023webarena}) and it seems to be harder to identify the attribution of errors due to the complex nature of real websites.
\update{Moreover, while realistic benchmarks~\citep{deng2023mind2web,zhou2023webarena} often provide deterministic instructions and observations, CompWoB has randomized instructions generated from a set of templates, as well as element-randomized HTML.}
For the task complexity compared to other benchmarks, we provide the statistics (mean, max, percentiles) of episode steps and instruction tokens from MiniWoB~\citep{shi2017miniwob}, Mind2Web, and CompWoB in \autoref{tab:benchmark_comparison}. In terms of episode length and instruction length, CompWoB successfully increases task complexity from MiniWoB, and has sufficient difficulties the same as Mind2Web, a representative benchmark from real websites.

\begin{figure*}[t]
\centering
\includegraphics[width=0.95\linewidth]{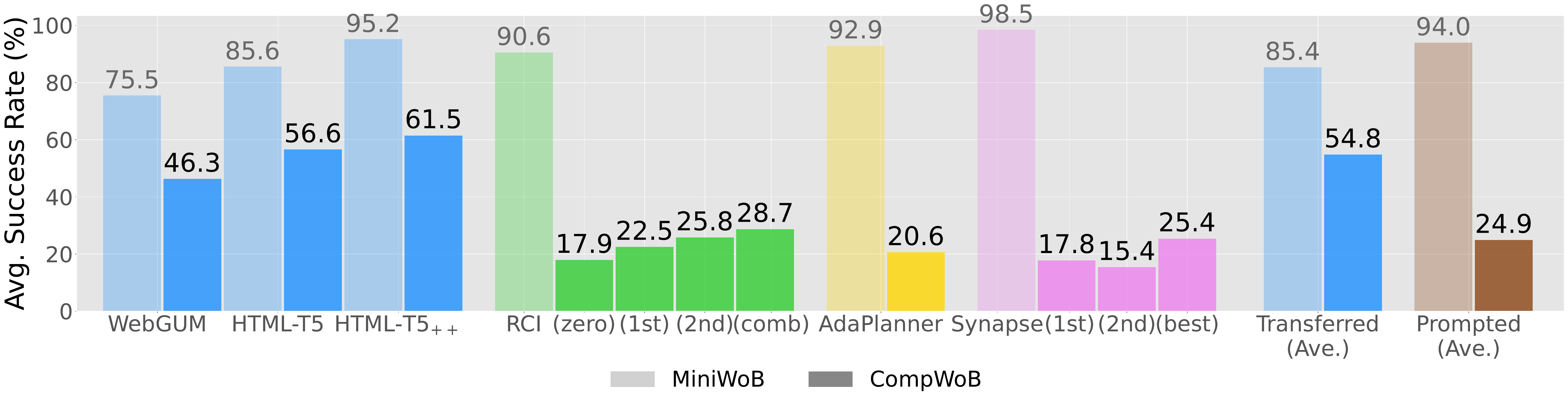}
\vskip -0.1in
\caption{
Average success rate of LMAs in 50 \compwob{} tasks. The light color represents the performance in the original MiniWoB, and the dark color for \compwob{}. We use \texttt{gpt-3.5-turbo} as the backbone LLM for prompted LMAs (RCI~\citep{kim2023rci}, AdaPlanner~\citep{sun2023adaplanner}, Synapse~\citep{zheng2023synapse}), and transferred LMAs with 3-billion parameters.
Transferred LMA, especially HTML-T5++, achieves the best generalization in compositional tasks, suppressing the performance degradation (from 95.2\% to 61.5\%).
On the contrary, prompted LMAs drop their performance significantly; even the best RCI that uses combined task prompts in the composition just achieves 28.7\% success (from 90.6\% in base tasks).
This indicates, in contrast to the base task performances, prompted LMAs are more vulnerable to, and transferred LMAs can deal with unknown task compositionality better than expected.
}
\vskip -0.2in
\label{fig:comp_success_rate}
\end{figure*}

\section{Results}
\label{sec:results}
\vskip -0.1in
\textbf{Evaluation Methodology}~
We evaluate both transferred and prompted LMAs with base MiniWoB demonstrations on the unseen compositional tasks in a ``\textbf{zero-shot}'' manner; i.e. \textbf{we do not provide any demonstrations on the compositional tasks for the training corpus and exemplars to measure the generalization}.
We test 50 compositional tasks and run 100 episodes per task.
We adopt \texttt{gpt-3.5-turbo} as a backbone LLM, unless otherwise mentioned.
We assume the optimal exemplar retriever throughout experiments and always provide the pre-defined prompts to LMAs.
We borrow hyper-parameters and prompt templates from respective papers with minimal change to respect our zero-shot transfer setting.
As a sanity check for the quality of environments and baseline agents, we also test LMAs with oracle demonstrations in \autoref{appendix:oracle}.

\textbf{RCI}~
We test 4 prompting strategies: (1) zero-shot (without any exemplars), few-shot with (2) first-task exemplars, (3) second-task exemplars, and (4) combination exemplars (i.e. both first and second tasks).
For consistency and limited context length, we always consider the first two tasks even if the number of primitive tasks is more than two, and fix the number of self-improvement iterations to 1 explicit RCI and 3 implicit RCI as recommended in the original paper.
The exemplars we use are provided by \citet{kim2023rci}.

\textbf{AdaPlanner}~
Following the original code, we use the same exemplars provided by \citet{sun2023adaplanner} for tasks where those base tasks are included, such as \texttt{enter-text} and \texttt{click-widget} (see Appendix~\ref{sec:hyperparams_adaplanner}).
Otherwise, the agents are prompted in a zero-shot manner.

\textbf{Synapse}~
We test 3 prompting strategies: few-shot with (1) first-task, (2) second-task exemplars, and (3) best exemplars (i.e. maximum score between (1) and (2)).
Because prompts and modules are quite different among tasks, we do not merge the prompts and use proper hyper-parameters corresponding to the given exemplars designed by \citet{zheng2023synapse}.

\begin{table}[b]
\begin{center}
\begin{small}
\scalebox{0.525}{
\begin{tabular}{l|l|l|l|l|l}
\toprule
\multicolumn{2}{c|}{\textbf{RCI}~\citep{kim2023rci}} & \multicolumn{2}{c|}{\textbf{AdaPlanner}~\citep{sun2023adaplanner}} & \multicolumn{2}{c}{\textbf{Synapse}~\citep{zheng2023synapse}} \\
\midrule
\multicolumn{2}{l|}{\textit{Click button ONE, then click button TWO, and then select whX, 1Nk,}} & \multicolumn{2}{l|}{\textit{Enter the password "UBKR" into both text fields, and then}} & \multicolumn{2}{l}{\textit{Select yE, and then enter "Juan" into the text field and}} \\
\multicolumn{2}{l|}{\textit{fUK3 and click Submit}} & \multicolumn{2}{l|}{\textit{select KwpUv and click Submit}} & \multicolumn{2}{l}{\textit{press Submit}} \\
\midrule
 \multicolumn{1}{c|}{\textcolor{cb_green}{\CheckmarkBold}} & \multicolumn{1}{c|}{\textcolor{cb_red}{\XSolidBrush}} & \multicolumn{1}{c|}{\textcolor{cb_green}{\CheckmarkBold}} & \multicolumn{1}{c|}{\textcolor{cb_red}{\XSolidBrush}} & \multicolumn{1}{c|}{\textcolor{cb_green}{\CheckmarkBold}} & \multicolumn{1}{c}{\textcolor{cb_red}{\XSolidBrush}} \\
\textcolor{cb_green}{1. click //button[@id="subbtn1"]} &\textcolor{cb_red}{1. type //button[@id="subbtn2"]} & \textcolor{cb_green}{1. click //*[@id="password"]} &  & \textcolor{cb_green}{1. click //*[text()="yE"]/input} &  \\
\textcolor{cb_green}{2. click //button[@id="subbtn2"]} &2. click //*[text()="whX"]/input  & 2. type UBKR & 1. type UBKR & 2. click //input[@id="tt"] & 1. click //input[@id="tt"] \\
3. click //*[text()="whX"]/input & 3. click //*[text()="1Nk"]/input & \textcolor{cb_green}{3. click //*[@id="verify"]} &  &  & \textcolor{cb_red}{2. type yE} \\
4. click //*[text()="1Nk"]/input  & 4. click //*[text()="fUK3"]/input & 4. type UBKR & 2. type UBKR & 3. type Juan &  3. type Juan \\
5. click //*[text()="fUK3"]/input  & \textcolor{cb_red}{5. click //*[text()="gSm"]/input} & 5. click //input[@id="ch0"] & 3. click //input[@id="ch0"] & 4. click //*[@id="subbtn"] & 4. click //*[@id="subbtn"] \\
6. click //*[@id="subbtn"] & 6. click //*[@id="subbtn"] & 6. click //*[@id="subbtn"] & 4. click //*[@id="subbtn"] &  &  \\
\bottomrule
\end{tabular}
}
\end{small}
\end{center}
\vskip -0.15in
\caption{
Failure examples in \compwob{}. The left columns have correct plans and the right have failure plans.
LMAs often ignore necessary intermediate steps or predict incorrect action types and XPath.
}
\label{tab:error_comp_miniwob}
\end{table}

\subsection{Language Model Agents Struggle to Handle Task Compositionality}
\label{sec:compositional_main}
\autoref{fig:comp_success_rate} shows that, in \compwob{}, all the LMAs face performance degradation.
Among those, transferred LMAs achieve better success rate (54.8\%) than prompted LMAs (24.9\%) on average.
In particular, HTML-T5++ achieves the best generalization, suppressing the performance drop from 95.2\% to 61.5\%.
\update{This might be because finetuned LLMs are better at natural language compositional tasks prompted LLMs in general. See Appendix~\ref{appendix:lastletter} for further details.}
In contrast, prompted LMAs degrade their performance drastically; even the best RCI with few-shot combination exemplars (comb) just degrades the success rate to 28.7\% from 90.6\% in base MiniWoB.
These results indicate that LMAs suffer from generalization to task compositionality, and transferred LMAs can relatively deal with that better than prompted LMAs, which is an opposite trend to base MiniWoB performance.
Among prompted LMAs, RCI performs better than AdaPlanner and Synapse, which suggests that multiple iterations of self-criticism and improvement might be more robust to out-of-domain decision making from the exemplars than program synthesis with feedback or structured prompting with state translations.

In the failure episodes (\autoref{tab:error_comp_miniwob}), LMAs often miss necessary steps, common to all the prompted LMAs. Since the instructions get long in compositional settings, LMAs may skip important intermediate steps to satisfy the instructions.
In addition, they predict incorrect action types and XPath: for instance, hallucination in XPath (RCI) and mixing up \textit{click} and \textit{type} action (Synapse).
\update{Without oracle exemplars, LMAs often struggle to parse compositional task instructions and understand complex HTML.}
\autoref{sec:per_task} also provides average success rates per category.

\begin{figure*}[t]
\centering
\includegraphics[width=0.95\linewidth]{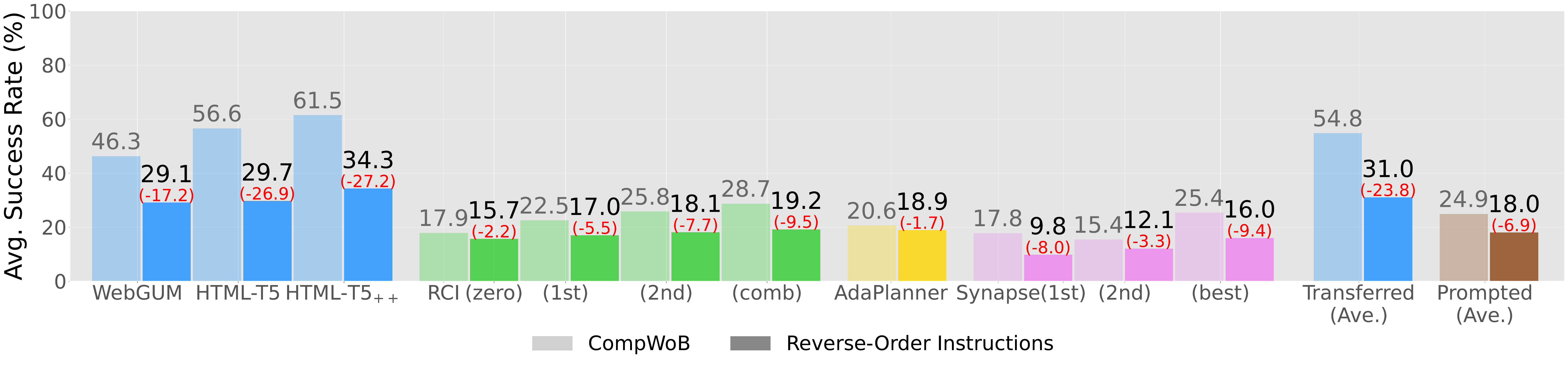}
\vskip -0.1in
\caption{
Average success rate of language model agents in \textbf{reverse-order instruction} settings.
We use \texttt{gpt-3.5-turbo} as the backbone LLM for prompted LMAs (RCI, AdaPlanner, Synapse), and transferred LMAs with 3-billion parameters.
Notably, most LMAs significantly degrade the success rate when reverse-order instructions are provided (the performance gap is highlighted with a \textcolor{cb_red}{red} number).
This trend is more remarkable in transferred LMA (54.8\% $\rightarrow$ 31.0\% on average) than prompted LMA (24.9\% $\rightarrow$ 18.0\% on average), which suggests that any kind of LMAs are susceptible to the order of compositional instructions. The capability as general language models might be important to parse semantically complex instructions into the correct plan.
}
\vskip -0.2in
\label{fig:reverse_success_rate}
\end{figure*}

\subsection{Reverse-Order Instructions Degrade Language Model Agents}
\label{sec:reverse_order}
As shown in \autoref{fig:reverse_success_rate}, all the LMAs significantly degrade the success rate when reverse-order instructions are provided.
This trend is more remarkable in transferred LMAs dropping from 54.8\% to 31.0\% than prompted LMAs from 24.9\% to 18.0\% on average, which suggests that any kind of LMAs is susceptible to the order of compositional instructions and that transferred LMAs may not generalize well to diverse instructions beyond the dataset distribution.
As opposed to \Cref{sec:compositional_main}, the performance differences among prompted LMAs are marginal, which, however, implies that existing prompting methods, even with self-improvement, may not handle complex task instructions enough.
The stronger capability as general-purpose language models or other prompting methods might be important to parse semantically complex instructions into the executable sequential plan.

Compared to Section~\ref{sec:compositional_main}, RCI and Synapse cannot parse reverse-order instructions into plans correctly (\autoref{tab:error_reverse_instruction}).
This can be because base-task exemplars in the prompt just have ``left-to-right'' instructions from base MiniWoB tasks; strongly conditioning LMAs to process instructions in a linear order while compositionality could imply a non-linear processing of instructions.
The mismatch between the prompt and tasks could cause the performance drops.
LMAs still fail to select correct action types (AdaPlanner) or XPath (Synapse), and they also predict unnecessary actions (RCI).

\begin{table}[b]
\begin{center}
\begin{small}
\scalebox{0.525}{
\begin{tabular}{l|l|l|l|l|l}
\toprule
\multicolumn{2}{c|}{\textbf{RCI}~\citep{kim2023rci}} & \multicolumn{2}{c|}{\textbf{AdaPlanner}~\citep{sun2023adaplanner}} & \multicolumn{2}{c}{\textbf{Synapse}~\citep{zheng2023synapse}} \\
\midrule
\multicolumn{2}{l|}{\textit{Select rJ and click Submit, after clicking on the "yes" button}} & \multicolumn{2}{l|}{\textit{Select OkRi7,  and click Submit, after clicking on the "previous" button}} & \multicolumn{2}{l}{Select 2ld1 and click Submit, after entering the password "Zy4XI"} \\
\multicolumn{2}{l|}{} & \multicolumn{2}{l|}{} & \multicolumn{2}{l}{\textit{into both text fields}} \\
\midrule
 \multicolumn{1}{c|}{\textcolor{cb_green}{\CheckmarkBold}} & \multicolumn{1}{c|}{\textcolor{cb_red}{\XSolidBrush}} & \multicolumn{1}{c|}{\textcolor{cb_green}{\CheckmarkBold}} & \multicolumn{1}{c|}{\textcolor{cb_red}{\XSolidBrush}} & \multicolumn{1}{c|}{\textcolor{cb_green}{\CheckmarkBold}} & \multicolumn{1}{c}{\textcolor{cb_red}{\XSolidBrush}} \\
\textcolor{cb_green}{1. click //button[text()="yes"]}  & \textcolor{cb_red}{1. click //*[text()="rj"]/input} & 1. click //*[text()="previous"] & 1. click //*[text()="previous"] & \textcolor{cb_green}{1. click //*[@type="password"]} &  \textcolor{cb_red}{1. click //*[text()="2ld1"]/input} \\
\textcolor{cb_green}{2. click //*[text()="rj"]/input} & \textcolor{cb_red}{2. click //button[text()="yes"]} & \textcolor{cb_green}{2. click //*[text()="OkRi7"]/input} & \textcolor{cb_red}{2. type OkRi7} & 2. type Zy4XI & \textcolor{cb_red}{2. click //*[@type="password"][1]} \\
& \textcolor{cb_red}{3. type rj} & 3. click //*[@id="subbtn"] &  3. click //*[@id="subbtn"] & \textcolor{cb_green}{3. click //*[text()="verify"]} & 3. type Zy4XI \\
3. click //*[@id="subbtn"]  & 4. click //*[@id="subbtn"] &  &  & 4. type Zy4XI & \textcolor{cb_red}{4. click //*[@type="password"][2]} \\
  &  &  &  & \textcolor{cb_green}{5. click //*[text()="2ld1"]/input} & 5. type Zy4XI \\
  &  &  &  & 6. click //*[@id="subbtn"] & 6. click //*[@id="subbtn"] \\
\bottomrule
\end{tabular}
}
\end{small}
\end{center}
\vskip -0.15in
\caption{
Failure examples in \compwob{} with \textbf{reverse-order} instructions.
LMAs often fail to parse the instruction into the correct-order plan, and hallucinate unnecessary actions (e.g. \textit{type}). 
}
\label{tab:error_reverse_instruction}
\end{table}

\begin{figure*}[t]
\centering
\includegraphics[width=0.9\linewidth]{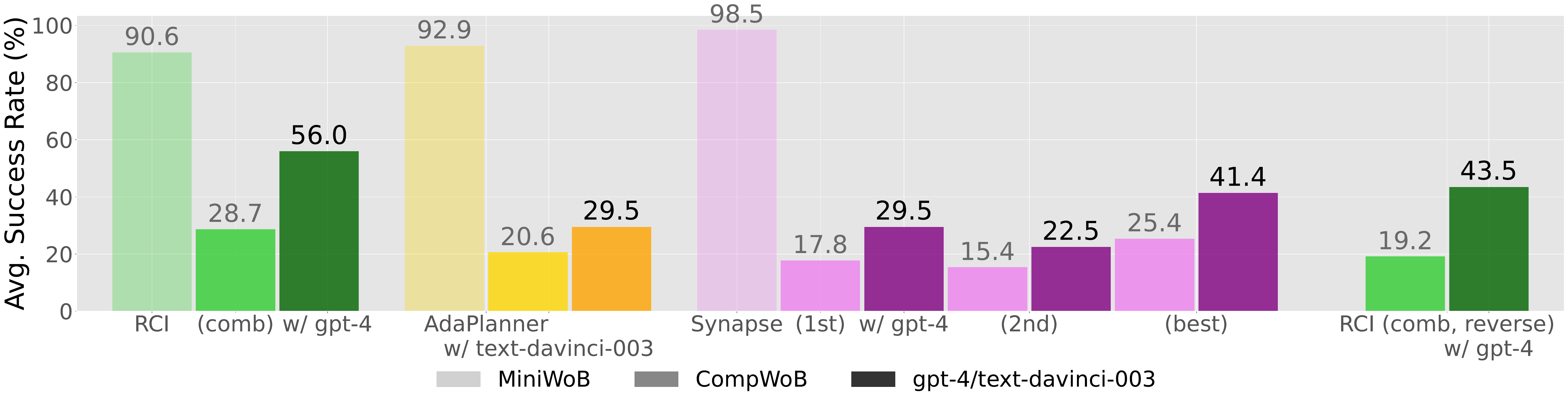}
\vskip -0.1in
\caption{
Average success rate of large language models with advanced LLMs (\texttt{gpt-4} for RCI and Synapse, and \texttt{text-davinci-003} for AdaPlanner).
The lighter color represents the performance in MiniWoB, the medium color does in \compwob{} with \texttt{gpt-3.5-turbo}, and the darker color does with \texttt{gpt-4} or \texttt{text-davinci-003}.
The more capable models (\texttt{gpt-4} as a generalist and \texttt{text-davinci-003} as a coder) can improve the success rate of prompted LMAs but still struggle to generalize to compositional tasks (e.g. 56.0\% by RCI) or to deal with reverse-order instructions  (e.g. 43.5\% by RCI).
This may indicate that we need much better LLMs to realize deployable agents in the complex real world.
}
\vskip -0.2in
\label{fig:model_ablation}
\end{figure*}

\subsection{Do Advanced LLMs Solve Compositional Tasks?}
\label{sec:gpt_4}
\autoref{fig:model_ablation} presents the results when we adopt other advanced LLMs, than \texttt{gpt-3.5-turbo}, as a backbone of each LMA.
The more capable models, such as \texttt{gpt-4} in a generalist aspect and \texttt{text-davinci-003} in a code generation, can improve the success rate of all the prompted LMAs.
However, even \texttt{gpt-4} is still far from the generalization in compositional tasks (from 28.7\% to 56.0\% by RCI) or from dealing with reverse-order instructions (from 19.2\% to 43.5\% by RCI).
This indicates that we need much better LLMs to realize deployable systems in complex real-world decision-making tasks.
We provide failure examples in \autoref{sec:error_gpt_4}.

\subsection{What Determines Task Complexity on the Web?}
\label{sec:task_complexity_main}
\begin{figure*}[t]
\centering
\includegraphics[width=0.9\linewidth]{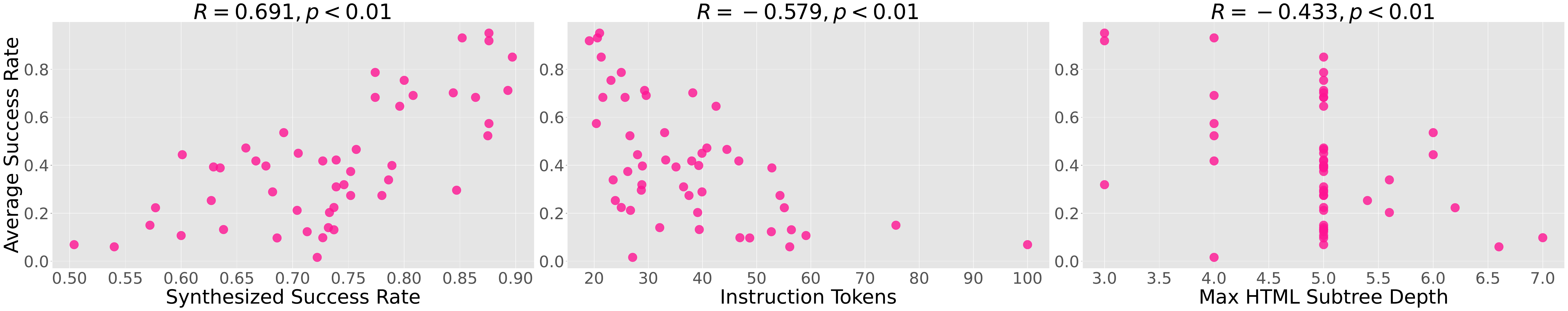}
\vskip -0.1in
\caption{
2D-scatter plots between success rate averaged among LMAs (y-axis) and each statistic of compositional task (x-axis), such as success rate synthesized with a product of base task success rate, the number of instruction tokens, and max depth of HTML subtrees.
Synthesized success rate positively correlates with an average success rate ($R=0.691$, statistically significant in paired t-test with $p<0.01$), indicating that base task difficulty may determine compositional task difficulty. In addition, the number of instruction tokens ($R=-0.579$; $p<0.01$) and the max depth of HTML subtrees ($R=-0.433$; $p<0.01$) show negative correlations, which suggests the high complexity of observation and long instructions make the compositional tasks hard to resolve.
}
\vskip -0.15in
\label{fig:task_complexity}
\end{figure*}

\autoref{fig:task_complexity} visualizes the correlation between the success rate averaged across WebGUM, HTML-T5, RCI, AdaPlanner, and  Synapse (y-axis) and each statistic of compositional tasks (x-axis), such as synthesized success rate -- a product of base task success rates among compositional tasks -- the number of instruction tokens, and max depth of HTML subtrees.
Synthesized success rate positively correlates with an average success rate ($R=0.691$), indicating that compositional task difficulty takes over base task difficulties. In addition, the number of instruction tokens ($R=-0.579$) and the max depth of HTML subtrees ($R=-0.433$) show negative correlations. All those are statistically significant in paired t-test with $p<0.01$.
In contrast, other task statistics, such as synthesized success rate with human performance, the number of HTML tokens, and elements in HTML, just show relatively weaker correlations (see \autoref{sec:task_complexity_full} for the details).
This analysis suggests that HTML with larger depth and long instructions make generalizing compositional tasks challenging.
The complexity of HTML is determined by its depth rather than its length or the number of elements. This might come from the hierarchical nature of HTML; in deeper HTML subtrees, the elements near the root tend to be distant from each other after the traversal. Such sparsity may cause confusion during planning.
We also report the individual characteristics of each agent in \autoref{appendix:per_method_complexity}.

\section{Discussion}
\textbf{Generalizable Prompting Methods}~
The results of Synapse and RCI in \autoref{fig:comp_success_rate} imply that those prompted LMAs have ``over-fitting'' trends to the base MiniWoB tasks.
While the robustness across the prompts has been investigated in natural language tasks~\citep{wei2022cot,kojima2022lets}, it is not well understood in decision making.
Because we will not be able to prepare the optimal self-improvement iterations or decomposed prompts for all the possible instructions and task compositions, even if using optimal exemplar retrievers, we should care more about the generalization of prompting methods for the agent systems.

\textbf{Agent-Specialized Large Language Models}~
As shown in \autoref{fig:model_ablation}, the more capable LLMs, such as \texttt{gpt-4}, can improve the performance of LMAs in \compwob{}. However, it has not reached the base MiniWoB yet (e.g. from 90.6\% to 56.0\% in RCI, and from 98.5\% to 41.4\% in Synapse).
Similarly, as described in Section~\ref{sec:finetuned_lma}, transferred LMAs can perform better if the training dataset has a good balance and coverage, but it is far from sufficient generalization to compositionality and instructions.
The current pre-trained LLMs may still not be sufficient to generalize to complex decision making tasks, and then, in addition to prompting methods, the development of agent-specialized LLMs with enhanced reasoning and generalization through the instruction-tuning~\citep{zeng2023agenttuning} or RLHF/AIF~\citep{furuta2024geometric} would be expected.


\textbf{Parsing Complex Instructions to Executable Plan}~
Section~\ref{sec:reverse_order} highlights that LMAs are fragile when we increase the complexity of instruction even by the most straightforward reverse-order instructions.
This may not be preferable for the real-world application since the instructions might not be easy-to-parse and the users should carefully and concisely tell what they would like to do, which hinders the user's experience.
It would be an interesting future direction to investigate better planning modules that could parse complex instructions to correct and executable plans.

\section{Conclusion}
The robustness and generalization of LMAs are important aspects for real-world deployment.
We extensively examine how much existing LMAs, via transferring and prompting, can deal with a set of compositional web automation environments, \compwob{}, that consists of easily-resolvable base primitive tasks.
Our evaluation implies the contrary conclusion to the prior works (\autoref{tab:summary_table});  the prompted LMAs are strong solver for primitive web automation tasks but significantly drop their performance in unknown task compositionality. The transferred LMAs often show sub-optimal performance in basic tasks but can deal with compositional problems much better.
Our detailed analysis also highlights that LMAs also face catastrophic degradation when they receive complex, even in the simplest reversed-order instructions, and that the challenges in compositional tasks might come from instruction length and the depth of HTML subtree.
We hope this inspires the community to build robust and generalizable LMAs to task compositionality toward real-world application.

\begin{table}[thb]
\begin{center}
\begin{small}
\scalebox{1.0}{
\begin{tabular}{lcccc}
\toprule
& Base  &  & Reverse-Order & Advanced \\
& MiniWoB & \compwob{} & Instructions & Models \\
\midrule
Prompted LMA & \textbf{94.0}\% / \textbf{98.5}\% & 24.9\% / 28.7\% & 18.0\% / 19.2\% & 42.3\% / 56.0\% \\
Transferred LMA & 85.4\% / 95.2\% & \textbf{54.8}\% / \textbf{61.5}\% & \textbf{31.0}\% / \textbf{34.3}\% & -- \\
\bottomrule
\end{tabular}
}
\end{small}
\end{center}
\vskip -0.15in
\caption{
Summary of average / max success rate in web automation.
}
\vskip -0.1in
\label{tab:summary_table}
\end{table}

\subsubsection*{Ethics Statements}
This paper systematically evaluates the performance of existing language model agents for web automation in realistic compositional tasks, and unveils remaining challenges towards broader generalization and real-world deployments. This technique could realize capable AI assistant tools on digital devices (e.g. computers or smartphones), and improve productivity and accessibility for society.

While we anticipate the positive aspects of autonomous agents, for responsible development, we should also consider the potential harmful applications and unintended consequences.
The misuse of web automation would threaten cyber security, and the users may get scammed. To reduce these risks, it is essential for researchers, policymakers, and industry to discuss concrete guidelines and regulations.

\subsubsection*{Acknowledgments}
We thank Mustafa Safdari, Yingjie Miao, Yusuke Iwasawa, Heiga Zen, and Doina Precup for the help on this work. HF was supported by JSPS KAKENHI Grant Number JP22J21582.

\bibliography{colm2024_conference}
\bibliographystyle{tmlr}

\clearpage
\appendix
\section*{Appendix}
\section{Details of LLM API}
We used OpenAI API to call LLM inference in our experiments. \autoref{tab:llm_api_list} shows the API used for each method. We did most of our experiments from 2023/07 to 2023/09.
We use the official implementations and prompts released by the authors~\footnote{\url{https://github.com/posgnu/rci-agent}}\footnote{\url{https://github.com/haotiansun14/AdaPlanner}}\footnote{\url{https://github.com/ltzheng/Synapse}}.
We spent about \$3.6K for the experiments in total.

\begin{table}[ht]
\begin{center}
\begin{small}
\scalebox{0.925}{
\begin{tabular}{llll}
\toprule
\textbf{Methods} & \textbf{API} & \textbf{Cost} (input/output; /1K tokens) & \textbf{Context Length} \\
\midrule
RCI~\citep{kim2023rci} & \texttt{\texttt{gpt-3.5-turbo}} & \$0.0015 / \$0.002 & 4K tokens\\
 & \texttt{gpt-4} & \$0.03 / \$0.06 & 8K tokens\\
AdaPlanner~\citep{sun2023adaplanner} & \texttt{gpt-3.5-turbo} & \$0.0015 / \$0.002 & 4K tokens\\
 & \texttt{text-davinci-003} & \$0.02 / \$0.02 & 4K tokens\\
Synapse~\citep{zheng2023synapse} & \texttt{gpt-3.5-turbo} & \$0.0015 / \$0.002 & 4K tokens\\
 & \texttt{gpt-4} & \$0.03 / \$0.06 & 8K tokens\\
\bottomrule
\end{tabular}
}
\end{small}
\end{center}
\vskip -0.15in
\caption{List of LLM API used in this paper. We did those experiments from 2023/07 to 2023/09.}
\label{tab:llm_api_list}
\end{table}

\section{Pseudo Code for Prompted Language Model Agents}
\label{sec:lma_pseudo_code}
We explicitly distinguish LLM itself and LMA (combination of LLM, prompting and pipeline) in this paper. We provide the pseudo code for prompted LMAs to clarify their methodological difference.
RCI~\citep{kim2023rci} is the first to use prompting in a self-refinement loop, outperforming imitation- or reinforcement-learned agents on MiniWoB benchmark that requires millions of demonstrations to work. AdaPlanner~\citep{sun2023adaplanner} and Synapse~\citep{zheng2023synapse} are the follow-up works outperforming RCI via code generation from environmental feedback or via well-designed decomposed prompts with retrieval.

\begin{algorithm}[th]
\caption{Prompted Language Model Agents: \textcolor{cb_green}{RCI}, \textcolor{cb_orange}{AdaPlanner}, \textcolor{cb_purple}{Synapse}}
\renewcommand{\algorithmicrequire}{\textbf{Input:}}
\label{alg:lma}
\begin{algorithmic}[1]

\Require prompt $P$, LMA $\pi$, task $\psi$, environment  \texttt{Env}, large language model \texttt{LLM}, \textcolor{cb_green}{number of ERCI $N_{\text{ERCI}}$}, \textcolor{cb_green}{number of IRCI $N_{\text{IRCI}}$}
\State $s, g \xleftarrow{} \texttt{Env.reset($\psi$)}$
\State \textcolor{cb_purple}{$s, g \xleftarrow{} \texttt{LLM}(\cdot|P_{\text{syn}}, s, g)$} \Comment{\textcolor{cb_purple}{Task Reformulation (Synapse)}}
\State{$\texttt{history} \xleftarrow{} \{\}$}
\While{\texttt{Env} is not terminated}
    \State{$\{a_1, ..., a_T\} \xleftarrow{} \pi(\cdot|P_{\pi}, s, g)$}  \Comment{Planning}
    \For{$i ~~\textbf{in}~~\texttt{range(}\textcolor{cb_green}{N_{\text{ERCI}}}\texttt{)}$}
        \State{\textcolor{cb_green}{$\texttt{criticism} \xleftarrow{} \texttt{LLM}(\cdot|P_{\text{rci}}, \{a_1, ..., a_T\})$}}  \Comment{\textcolor{cb_green}{Criticism (RCI)}}
        \State{\textcolor{cb_green}{$\{a_1, ..., a_T\} \xleftarrow{} \pi(\cdot|P_{\pi}, s, g, \{a_1, ..., a_T\},\texttt{criticism})$}}  \Comment{\textcolor{cb_green}{Improvement (RCI)}}
    \EndFor
    \For{$a ~~\textbf{in}~~\{a_1, ..., a_T\}$}
        \For{$j ~~\textbf{in}~~\texttt{range(}\textcolor{cb_green}{N_{\text{IRCI}}}\texttt{)}$}
            \State{\textcolor{cb_green}{$a \xleftarrow{} \pi(\cdot|P_{\pi}, s, g, \{a, ..., a_T\},\texttt{history})$}}  \Comment{\textcolor{cb_green}{Improvement (RCI)}}
        \EndFor
        \State{$s, r, \text{info} \xleftarrow{} \texttt{Env.step(a)}$}
        \State{\textcolor{cb_orange}{$\{a, ..., a_T\} \xleftarrow{} \pi(\cdot|P_{\pi}, s, g, \{a, ..., a_T\}, \texttt{history}, \text{info})$}}  \Comment{\textcolor{cb_orange}{Replanning (AdaPlanner)}}
        \State{$\texttt{history} \xleftarrow{} \texttt{history} \cup \{a\}$}
    \EndFor
\EndWhile
\end{algorithmic}
\end{algorithm}

\clearpage
\section{Details of Hyper-parameters}

\subsection{RCI}
\label{sec:hyperparams_rci}
As we described in Section~\ref{sec:lma_rci}, RCI has two important hyper-parameters to control the number of self-improvement iterations.
In Explicit RCI (ERCI) loop, LLMs criticize their own generated plans to identify the problem and then improve it, reflecting self-criticism.
In Implicit RCI (IRCI) loop, LLMs ground the action to the current state (i.e. HTML) and refine its formatting to be parsable without self-criticism, which may reduce hallucinations or tiny errors.
We here test how many self-improvement loops RCI requires (IRCI: 1-4, ERCI: 0-2). \autoref{tab:rci_hyperparams} shows that the optimal number of inference loops is different among tasks, while the recommendations are $\text{ERCI}=1$ and $\text{IRCI}=3$.
These two hyper-parameters might need to be adjusted for each task.

\begin{table}[hb]
\begin{center}
\begin{small}
\scalebox{0.8}{
\begin{tabular}{lrrrrrrrrrrrr}
\toprule
 & \multicolumn{12}{c}{\textbf{(ERCI, IRCI)}} \\
\cmidrule(r){2-13}
\textbf{Tasks} & (0,1) & (0,2) & (0,3) & (0,4) & (1,1) & (1,2) & \textbf{(1,3)} & (1,4) & (2,1) & (2,2) & (2,3) & (2,4) \\
\midrule
\texttt{click-button} & \textbf{1.00} & \textbf{1.00} & \textbf{1.00} & \textbf{1.00} & 0.92 & 0.84 & 0.87 & 0.88 & 0.93 & 0.86 & 0.87 & 0.87 \\
\texttt{click-checkboxes} & 0.90 & 0.94 & 0.87 & 0.91 & 0.96 & 0.94 & 0.97 & \textbf{1.00} & 0.89 & 0.91 & 0.94 & 0.99 \\
\texttt{click-dialog} & \textbf{1.00} & \textbf{1.00} & \textbf{1.00} & \textbf{1.00} & \textbf{1.00} & \textbf{1.00} & \textbf{1.00} & \textbf{1.00} & \textbf{1.00} & \textbf{1.00} & \textbf{1.00} & \textbf{1.00} \\
\texttt{click-link} & 0.96 & 0.95 & 0.98 & \textbf{0.99} & 0.91 & 0.91 & 0.89 & 0.88 & 0.95 & 0.91 & 0.91 & 0.87 \\
\texttt{click-option} & 0.82 & 0.77 & 0.79 & \textbf{0.87} & 0.41 & 0.54 & 0.56 & 0.52 & 0.83 & 0.82 & 0.73 & \textbf{0.87}\\
\texttt{click-scroll-list} & \textbf{0.86} & 0.86 & 0.84 & 0.83 & 0.75 & 0.81 & 0.78 & 0.79 & 0.76 & 0.81 & 0.85 & 0.74\\
\bottomrule
\end{tabular}
}
\end{small}
\end{center}
\vskip -0.15in
\caption{The success rate of RCI with different hyper-parameters. The optimal parameters differ in each task, while the recommended one is (1,3).}
\label{tab:rci_hyperparams}
\end{table}

\subsection{AdaPlanner}
\label{sec:hyperparams_adaplanner}
We use the demonstrations of these 13 tasks where they are included in the task composition:
\footnotesize
\begin{itemize}[leftmargin=0.5cm,topsep=0pt,itemsep=0pt]
    \item \texttt{enter-text}
    \item \texttt{click-widget}
    \item \texttt{navigate-tree}
    \item \texttt{login-user-popup}
    \item \texttt{email-inbox-forward-nl-turk}
    \item \texttt{click-checkboxes-large}
    \item \texttt{click-tab-2-hard}
    \item \texttt{click-dialog-2}
    \item \texttt{search-engine}
    \item \texttt{click-checkboxes-soft}
    \item \texttt{use-autocomplete}
    \item \texttt{enter-date}
    \item \texttt{click-dialog-2}
\end{itemize}
\normalsize

\subsection{Synapse}
\label{sec:hyperparams_synapse}
As we explained in Section~\ref{sec:lma_synapse}, Synapse has several hyper-parameters to construct optimal structured prompts per task to specify whether LLMs translate the instruction or HTML.

\autoref{tab:synapse_reformation} summarizes the type of reformulation into 7 categories and clarifies which transformed inputs are used for predicting open-loop plans. For instance, \texttt{Task} only requires translated instructions (and few-shot planning exemplars), although \texttt{Obs} takes raw instruction, HTML, and translated HTML as inputs.
For the tasks that require temporal abstraction, it also employs state-conditional decomposition, which factorizes demonstrations into a set of exemplars conditioned on the environmental states, and can reduce error accumulation over the time step.

\autoref{tab:hyperparams_synapse} provides the detailed values for state-filtering and task reformulation, which is quite different across the tasks.
These well-designed structured prompts could be the source of the best performance in base MiniWoB.
However, in compositional settings, it is challenging to modify them for any combinations. Instead, we assume the optimal retriever always picks up the exemplars for one of the base tasks, and we compute the maximum score among the results with given prompts.

\begin{table}[t]
\begin{center}
\begin{small}
\scalebox{0.75}{
\begin{tabular}{lccccccc}
\toprule
& \multicolumn{7}{c}{\textbf{Reformulation Strategies}} \\
\cmidrule(r){2-8}
\textbf{Inputs} & \texttt{Task} & \texttt{Obs} & \texttt{Obs\_Task} & \texttt{Obs\_Task\_Filter} & \texttt{Raw\_Task} & \texttt{None} & \texttt{None\_Filter}\\
\midrule
\textbf{Instruction}  & Translated & Raw & Translated & Translated & Raw & Raw & Raw \\
\textbf{HTML}  & \textcolor{cb_red}{\XSolidBrush} & Raw+Translated & Raw+Translated & Translated & \textcolor{cb_red}{\XSolidBrush} & Raw & Translated \\
\bottomrule
\end{tabular}
}
\end{small}
\end{center}
\vskip -0.15in
\caption{
Summary of task reformulation for structured prompting used in Synapse~\citep{zheng2023synapse}.
Structured prompts are finely designed per task.
}
\vskip -0.1in
\label{tab:synapse_reformation}
\end{table}

\begin{table}[hb]
\begin{center}
\begin{small}
\scalebox{0.725}{
\begin{tabular}{lrrrr}
\toprule
\textbf{Tasks} & \textbf{State Filtering} & \textbf{Exemplar Decomposition} & \textbf{Raw Task Only}  &  \textbf{Reformulation} \\
\midrule
book-flight & True & True & False & None \\
choose-date & False & False & False & Task \\
choose-list & False & False & True &  None \\
click-button & False & False & False & None \\
click-button-sequence & False & False & False & None \\
click-checkboxes & False & False & True & None \\
click-checkboxes-large & False & False &  True & None \\
click-checkboxes-soft & False & False & False & Obs \\
click-checkboxes-transfer & False & False & True & None \\
click-collapsible & False & False & True & None \\
click-collapsible-2 & False & False & False & Obs Task \\
click-color &False & False & True & None \\
click-dialog &False & False & True & None \\
click-dialog-2 &False & False & False &  None\\
click-link &False & False & True & None \\
click-menu &False & False & True & Task \\
click-option &False & False & True & None \\
click-pie &False & False & True & None \\
click-scroll-list &False & False & True & None \\
click-shades &False & False & True & Task \\
click-shape & True & False & False & Obs Task \\
click-tab &False & False & True & None \\
click-tab-2 & True & False & False & Obs Task \\
click-tab-2-hard & True & False & False & Obs Task \\
click-test &False & False & False & None \\
click-test-2 & False & False & False & None \\
click-widget &False & False & True & None \\
copy-paste &False & False & False & None \\
copy-paste-2 &False & False & False & None \\
count-shape & True  & False & False & Obs Task \\
email-inbox &False & False & False & Task \\
email-inbox-forward-nl  &False & False & False & Task \\
email-inbox-forward-nl-turk &False & False & False & Task \\
email-inbox-nl-turk &False & False & False & Task \\
enter-date &False & False & True & None \\
enter-password &False & False & True & None \\
enter-text & False& False & True & None \\
enter-text-dynamic &False & False & True & None \\
enter-time &False& False & True & None \\
find-word & True & False & False & None \\
focus-text & False& False & True & None \\
focus-text-2 & False& False & True & None \\
grid-coordinate & False& False & True & None \\
guess-number & False& False & False & None \\
identify-shape & False& False & False & None \\
login-user & False& False & True & None \\
login-user-popup & False & False & True & None \\
multi-layouts &False & False & False & None \\
multi-orderings &False & False & False & None \\
navigate-tree & False & False & False & None \\
read-table & False& False & False & None \\
search-engine & False & False & True & None \\
simple-algebra & False & False & False & None \\
simple-arithmetic & False & False & False & None \\
social-media & False & False & False & Obs Task \\
social-media-all & False & False & True & None \\
social-media-some &False & False & True & None \\
terminal & False & True & False & None \\
text-transform & False & False & False & None\\
tic-tac-toe & True & False & False & Obs Task \\
unicode-test &False & False & False & None\\
use-autocomplete &False & True & False & None  \\
use-spinner & False & False & False & Task \\

\bottomrule
\end{tabular}
}
\end{small}
\end{center}
\vskip -0.15in
\caption{Hyperparameters for Synapse~\citep{zheng2023synapse}. \textbf{Raw Task Only} is specified with \textbf{Task as Reformation} flag, and \textbf{Reformulation} is specified with \textbf{Reformat Input} flag in the original imprementation.
}
\label{tab:hyperparams_synapse}
\end{table}

\clearpage
\section{Ranking Base MiniWoB Tasks}
\label{sec:task_complexity}
To ensure the solvability of \compwob{} to some extent and to identify the data-redundant tasks for finetuned LMAs, we estimate the brute-force task difficulty~\citep{furuta2021pic} (\autoref{tab:miniwob_task_complexity}). We compute the average success rate for each task across representative previous web automation agents, such as CC-Net (SL, SL+RL)~\citep{humphreys2022data}, WGE~\citep{liu2018wge}, WebN-T5~\citep{gur2022html}, WebGUM~\citep{furuta2023mmwebnav}, HTML-T5~\citep{gur2023realworld}, RCI~\citep{kim2023rci}, AdaPlanner~\citep{sun2023adaplanner}, Pix2Act (BC, RL)~\citep{shaw2023pixels}, and Synapse~\citep{zheng2023synapse}. Based on those proxy difficulty measures, we classify 65 tasks into three categories~\citep{kim2023rci}: \texttt{easy} (from 0.8 to 1.0), \texttt{medium} (from 0.6 to 0.8), and \texttt{hard} (from 0.0 to 0.6).

\begin{table}[hb]
\begin{center}
\begin{small}
\scalebox{0.65}{
\begin{tabular}{llr}
\toprule
\textbf{Category} & \textbf{Task} & \textbf{Task Difficulty} \\
\midrule
\multirow{26}{*}{\texttt{easy} (0.8 - 1.0)} & click-button & 0.923 \\
 & click-button-sequence & 0.954 \\
 & click-checkboxes & 0.936 \\
 & click-checkboxes-transfer & 0.862 \\
 & click-collapsible & 0.878 \\
 & click-dialog & 0.923 \\
 & click-link & 0.949 \\
 & click-option & 0.839 \\
 & click-tab & 0.898  \\
 & click-test & 1.000 \\
 & click-test-2 & 0.996 \\
 & click-widget & 0.945 \\
 & email-inbox-forward-nl & 0.844 \\
 & email-inbox-forward-nl-turk & 0.804 \\
 & enter-password & 0.896 \\
 & enter-text & 0.922 \\
 & enter-text-dynamic & 0.933 \\
 & focus-text & 0.999 \\
 & focus-text-2 & 0.990 \\
 & grid-coordinate & 0.920 \\
 & identify-shape & 0.918 \\
 & login-user & 0.881 \\
 & login-user-popup & 0.818 \\
 & multi-layouts & 0.832 \\
 & navigate-tree & 0.825 \\
 & unicode-test & 0.900 \\
\midrule
\multirow{23}{*}{\texttt{medium} (0.6 - 0.8)} & choose-date-easy & 0.740 \\
 & click-checkboxes-large & 0.784 \\
 & click-checkboxes-soft & 0.754  \\
 & click-collapsible-2 & 0.693 \\
 & click-color & 0.742 \\
 & click-dialog-2 & 0.780 \\
 & click-menu & 0.607 \\
 & click-pie & 0.769 \\
 & click-shape  & 0.664 \\
 & click-tab-2 & 0.736 \\
 & click-tab-2-hard & 0.651 \\
 & copy-paste & 0.610 \\
 & email-inbox & 0.778 \\
 & email-inbox-nl-turk & 0.779 \\
 & enter-date & 0.714 \\
 & multi-orderings & 0.793 \\
 & read-table & 0.660 \\
 & search-engine & 0.723 \\
 & simple-algebra & 0.799 \\
 & simple-arithmetic & 0.782 \\
 & social-media & 0.733 \\
 & text-transform & 0.737 \\
 & use-autocomplete & 0.782 \\
\midrule
\multirow{16}{*}{\texttt{hard} (0.0 - 0.6)} & book-flight & 0.510 \\
 & choose-date & 0.331 \\
 & choose-date-medium & 0.497 \\
 & choose-list & 0.520 \\
 & click-scroll-list & 0.401 \\
 & click-shades & 0.501 \\
 & copy-paste-2 & 0.547 \\
 & count-shape & 0.536 \\
 & enter-time & 0.521 \\
 & find-word & 0.590 \\
 & guess-number & 0.363 \\
 & social-media-all & 0.432 \\
 & social-media-some & 0.532 \\
 & terminal & 0.592 \\
 & tic-tac-toe & 0.598 \\
 & use-spinner & 0.457 \\
\bottomrule
\end{tabular}
}
\end{small}
\end{center}
\vskip -0.1in
\caption{Brute-force task complexity and difficulty classification of MiniWoB. We split 65 tasks into the three category based on the task complexity: \texttt{easy} (0.8 - 1.0), \texttt{medium} (0.6 - 0.8), and \texttt{hard} (0.0 - 0.6).}
\label{tab:miniwob_task_complexity}
\end{table}

\clearpage
\section{Details of MiniWoB Dataset}
\label{sec:data_rebalance_details}
To resolve the data-imbalance problem, we first run Synapse~\citep{zheng2023synapse} on MiniWoB and collect 77K additional demonstrations across 16 tasks on top of 347K demonstrations~\citep{furuta2023mmwebnav} to compensate for the lack of data in specific tasks (\textbf{Strategy A}). We use PaLM 2-L~\citep{anil2023palm2} as a backbone LLM for Synapse.
We then reduce the number of demonstrations for the tasks the agents can solve to focus on more challenging tasks.
Based on brute-force task complexity~(\autoref{sec:task_complexity}), we consider the following four strategies as shown in \autoref{tab:dataset_ratio_details}:
\begin{itemize}[leftmargin=0.5cm,topsep=0pt,itemsep=0pt]
    \item Removing 50\% episodes from top-10 \texttt{easy} tasks (\textbf{Strategy B}; -73K)
    \item Removing 80\% episodes from top-10 \texttt{easy} tasks (\textbf{Strategy C}; -102K)
    \item Removing 50\% episodes from \texttt{easy} tasks (\textbf{Strategy D}; -142K)
    \item Removing 80\% episodes from top-15 \texttt{easy} tasks and removing 50\% episodes from other 11 \texttt{easy} tasks (\textbf{Strategy E}; -183K)
\end{itemize}
We hold out 21K episodes (5\% of 347K + 77K = 424K) as a validation split, and after the convergence, we choose the top-5 checkpoints that achieve higher offline validation accuracy, run those checkpoints online on MiniWoB benchmarks, and then report the best success rate.
Through the empirical evaluations (Section~\ref{sec:finetuned_lma}), we find that \textbf{Strategy D} realizes a well-balanced dataset to improve the performance.
Except for these dataset mixtures, we follow \citet{gur2023realworld} for other training hyper-parameters of HTML-T5++.
We have used cloud TPU-v3, which has a 32 GiB HBM memory space, with 128 cores. Each finetuning experiment takes about 1-2 days.

\begin{table}[ht]
\begin{center}
\begin{small}
\scalebox{0.75}{
\begin{tabular}{l|rrrrrr}
\toprule
\textbf{Task} & \textbf{Original} (347K) & \textbf{Strat. A} (424K) & \textbf{Strat. B} (351K) & \textbf{Strat. C} (322K) & \textbf{Strat. D} (282K) & \textbf{Strat. E} (241K) \\
\midrule
book-flight & 2.88\% & 2.49\% & 2.84\% & 3.11\% & 3.54\% & 4.15\% \\
choose-date &  0.11\% & 1.25\% & 1.42\% & 1.55\% & 1.77\% & 2.07\% \\
choose-date-easy &  0.97\% & 0.84\% & 0.95\% & 1.04\% & 1.19\% & 1.39\% \\
choose-date-medium &  0.64\% & 0.55\% & 0.63\% & 0.69\% & 0.79\% &  0.92\%\\
choose-list & 0.54\% & 1.24\% & 1.42\% & 1.55\% & 1.77\% & 2.07\% \\
click-button & 2.82\% & 2.44\% & 1.39\%& 0.61\%& 1.73\%& 0.81\% \\
click-button-sequence & 2.88\% & 2.49\% & 1.42\%& 0.62\%& 1.77\%& 0.83\% \\
click-checkboxes & 2.81\% & 2.43\% &  1.36\%&  0.55\%&  1.69\%& 0.73\% \\
click-checkboxes-large & 0.57\%  & 2.15\% & 2.46\% &  2.49\%&  3.06\%& 3.58\% \\
click-checkboxes-soft & 2.66\% & 2.30\% & 2.63\% &  2.87\%&  3.27\%& 3.83\% \\
click-checkboxes-transfer & 2.88\% & 2.49\% & 2.85\% &  3.11\%&  1.77\%& 2.07\% \\
click-collapsible  & 1.71\% & 1.48\% & 1.69\% &  1.85\%&  1.06\%& 1.24\% \\
click-collapsible-2 & 0.63\% & 0.55\% & 0.63\% &  0.68\%&  0.78\%&  0.91\% \\
click-color  & 0.74\% & 1.23\% & 1.40\% &  1.53\%&  1.74\%&  2.04\% \\
click-dialog  & 2.88\% & 2.49\% &  2.85\% &  3.11\%&  1.77\%& 0.83\% \\
click-dialog-2  & 0.95\% & 1.36\% & 1.55\% &  1.70\%&  1.93\%&  2.26\% \\
click-link & 2.87\% & 2.48\% & 1.42\% & 0.62\%&  1.76\%&  0.83\% \\
click-menu & 0.93\% & 1.24\% & 1.42\% &  1.55\%&  1.77\%& 2.07\% \\
click-option  & 2.88\% & 2.49\% & 2.84\% & 3.11\%&  1.77\%&  2.07\% \\
click-pie  & 1.07\% & 2.32\% & 2.65\% &  2.90\%&  3.30\%&  3.86\% \\
click-scroll-list & 0.00\% & 1.01\% & 1.16\% &  1.26\%&  1.44\%& 1.69\% \\
click-shades & 0.00\% & 1.25\% &  1.42\%&  1.55\%&  1.77\%&  2.07\% \\
click-shape  & 1.76\% & 1.80\% &  2.05\%&  2.24\%&  2.56\%& 2.99\% \\
click-tab  & 2.88\% & 2.49\% &  2.84\%&  3.10\%& 1.77\%& 2.07\% \\
click-tab-2 & 0.53\% & 0.46\% &  0.52\%&  0.57\%&  0.65\% & 0.76\% \\
click-tab-2-hard  & 0.45\% & 1.67\% & 1.91\%&  2.09\%& 2.38\% & 2.78\% \\
click-test & 2.88\% & 2.49\% & 1.42\%&  0.62\%& 1.77\% & 0.83\% \\
click-test-2  & 2.88\% & 2.49\% &  1.42\%&  0.62\%& 1.77\% & 0.83\% \\
click-widget  & 2.87\% & 2.48\% &  1.41\%&  0.62\%&  1.76\% & 0.82\% \\
count-shape  & 1.69\% & 1.10\% &  1.26\%&  1.37\%&  1.56\% &  1.83\% \\
email-inbox  & 1.49\% & 1.29\% &  1.47\%&  1.60\%&  1.83\% &  2.14\% \\
email-inbox-forward-nl  & 2.88\% & 2.49\% &  2.84\%&  3.11\%&  1.77\% & 2.07\% \\
email-inbox-forward-nl-turk & 1.41\% & 1.22\% &  1.39\%&  1.52\%& 0.86\% & 1.01\% \\
email-inbox-nl-turk  & 1.25\% & 1.08\% &  1.24\%&  1.35\%& 1.54\% & 1.80\% \\
enter-date  & 2.88\% & 2.49\% & 2.85\%&  3.11\%& 3.54\% & 4.15\% \\
enter-password  & 2.88\% & 2.49\% &  2.84\%&  3.10\%& 1.77\% & 2.07\% \\
enter-text  & 2.88\% & 2.49\% &  2.85\%&  3.11\%& 1.77\% & 0.83\% \\
enter-text-dynamic & 2.88\% & 2.49\% & 1.42\%&  0.62\%& 1.77\% & 0.83\% \\
enter-time & 0.00\% & 1.08\% &  1.23\%&  1.35\%& 1.54\% & 1.80\% \\
focus-text & 2.88\% & 2.49\% &  1.42\%&  0.62\%&  1.77\% &  0.83\% \\
focus-text-2  & 2.88\% & 2.49\% &  1.42\%&  0.62\%& 1.77\% & 0.83\% \\
grid-coordinate  & 2.41\% & 2.08\% &  2.38\%&  2.60\%& 1.41\% & 0.55\% \\
guess-number & 0.29\% & 1.25\% &  1.42\%&  1.55\%& 1.77\% & 2.07\% \\
identify-shape & 2.60\% & 2.24\% & 2.56\%&  2.80\%& 1.60\% & 0.75\% \\
login-user & 2.82\% & 2.44\% &  2.79\%&  3.05\%& 1.73\% & 2.03\% \\
login-user-popup  & 2.82\% & 2.44\% &  2.78\%&  3.04\%& 1.73\% & 2.03\% \\
multi-layouts & 2.88\% & 2.49\% &  2.85\%&  3.11\%&  1.77\% & 2.07\% \\
multi-orderings & 2.88\% & 2.49\% &  2.85\%&  3.11\%& 3.54\% & 4.15\% \\
navigate-tree  & 2.84\% & 2.46\% &  2.81\%&  3.07\%& 1.73\% &  2.02\% \\
search-engine  & 2.56\% & 2.21\% &  2.52\%&  2.76\%& 3.14\% & 3.68\% \\
social-media  & 0.76\& & 0.66\% &  0.75\%&  0.82\%& 0.93\% & 1.09\% \\
social-media-all & 0.03\% & 0.02\% &  0.03\%&  0.03\%&  0.03\% & 0.04\% \\
social-media-some & 0.09\& & 0.08\% &  0.09\%&  0.10\%& 0.11\% & 0.13\% \\
tic-tac-toe & 1.14\% & 1.43\% &  1.63\%&  1.78\%&  2.03\% & 2.37\% \\
use-autocomplete & 1.00\% & 0.86\% &  0.99\%&  1.08\%&  1.23\% & 1.44\% \\
use-spinner & 0.15\% & 1.19\% &  1.36\%&  1.48\%& 1.69\% & 1.98\% \\
\bottomrule
\end{tabular}
}
\end{small}
\end{center}
\vskip -0.1in
\caption{
Task ratio in rebalanced dataset for HTML-T5++.
}
\label{tab:dataset_ratio_details}
\end{table}

\clearpage
\section{Task Complexity Analysis in Web Automation}
\label{sec:task_complexity_full}
\begin{figure*}[h]
\centering
\includegraphics[width=\linewidth]{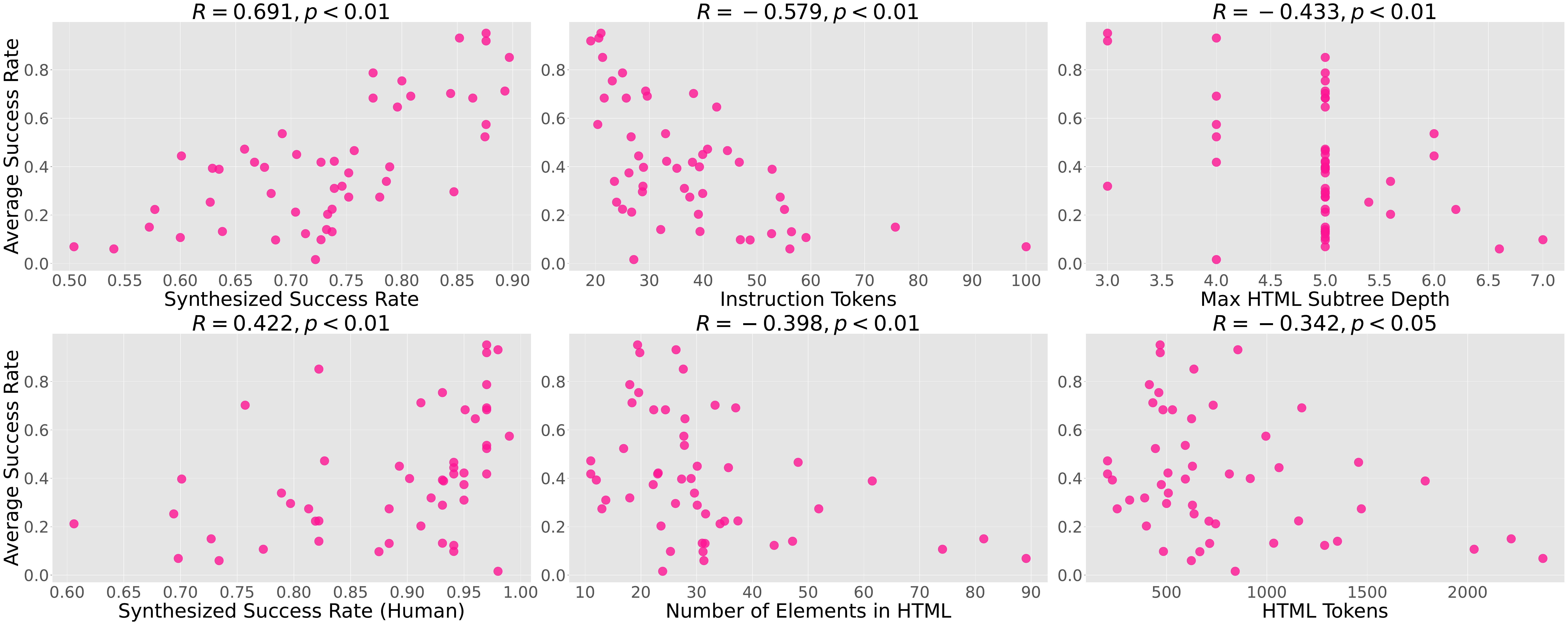}
\vskip -0.1in
\caption{
2D-scatter plots between success rate averaged among LMAs (y-axis) and each statistic of compositional task (x-axis), such as success rate synthesized with a product of base task success rate, the number of instruction tokens, max depth of HTML subtrees, success rate synthesized with a product of base task human performances, the number of elements in HTML, and the number of HTML tokens.
Synthesized success rate positively correlates with an average success rate ($R=0.691$, statistically significant in paired t-test with $p<0.01$), indicating that base task difficulty may determine compositional task difficulty. In addition, the number of instruction tokens ($R=-0.579$; $p<0.01$) and the max depth of HTML subtrees ($R=-0.433$; $p<0.01$) show negative correlations, which suggests the high complexity of observation and long instructions make the compositional tasks hard to resolve.
In contrast, synthesized success rate from human performance, the number of HTML tokens, and elements in HTML just show relatively weaker correlations.
}
\label{fig:task_complexity_full}
\end{figure*}

\section{Failure Examples with Advanced Language Models}
\label{sec:error_gpt_4}
We provide several failure episodes with advanced language models such as \texttt{gpt-4} and \texttt{text-davinci-003} in \autoref{tab:error_gpt4}. The left columns have correct plans, and the right columns have failure plans.
Qualitatively, LMAs based on advanced models manage to reduce the ratio of failure modes in \texttt{gpt-3.5-turbo} (\autoref{tab:error_comp_miniwob}), such as skipping necessary intermediate steps and hallucination of unnecessary actions.
However, they tend to suffer from tiny errors more such as capitalization (RCI, AdaPlanner) or attributes in HTML (Synapse).

\begin{landscape}
\begin{table}[hb]
\begin{center}
\begin{small}
\scalebox{0.74}{
\begin{tabular}{l|l|l|l|l|l}
\toprule
\multicolumn{2}{c|}{\textbf{RCI}~\citep{kim2023rci}} & \multicolumn{2}{c|}{\textbf{AdaPlanner}~\citep{sun2023adaplanner}} & \multicolumn{2}{c}{\textbf{Synapse}~\citep{zheng2023synapse}} \\
\midrule
\multicolumn{2}{l|}{\textit{Click on the link "adipiscing", and then click on the "submit" button}} & \multicolumn{2}{l|}{Click on the link "Augue", and then click on a "button" widget} & \multicolumn{2}{l}{Select 0Qm9EUt, and then enter the username "cristin" and the } \\
\multicolumn{2}{l|}{} & \multicolumn{2}{l|}{} & \multicolumn{2}{l}{password "M5" into the text fields and press login} \\
\midrule
 \multicolumn{1}{c|}{\textcolor{cb_green}{\CheckmarkBold}} & \multicolumn{1}{c|}{\textcolor{cb_red}{\XSolidBrush}} & \multicolumn{1}{c|}{\textcolor{cb_green}{\CheckmarkBold}} & \multicolumn{1}{c|}{\textcolor{cb_red}{\XSolidBrush}} & \multicolumn{1}{c|}{\textcolor{cb_green}{\CheckmarkBold}} & \multicolumn{1}{c}{\textcolor{cb_red}{\XSolidBrush}} \\
1. click //span[text()="adipiscing"]  & 1. click //span[text()="adipiscing"] & \textcolor{cb_green}{1. click //span[text()="Augue"]} & \textcolor{cb_red}{1. click //span[text()="augue"]} & \textcolor{cb_green}{1. click //*[text()="0Qm9EUt"]/input} &  \textcolor{cb_red}{1. click //*[@id="0Qm9EUt"]} \\
\textcolor{cb_green}{2. click //button[text()="submit"]}  & \textcolor{cb_red}{2. click //button[text()="Submit"]} & 2. click //*[@data-type="button"] & 2. click //*[@data-type="button"] & 2. click //*[@id="username"] & 2. click //*[@id="username"] \\
  &  &  &  & 3. type cristin & 3. type cristin \\
  &  &  &  & 4. click //*[@id="password"] & 4. click //*[@id="password"] \\
  &  &  &  & 5. type M5 & 5. type M5 \\
  &  &  &  & 6. click //*[@id="subbtn"] & 6. click //*[@id="subbtn"] \\
  &  &  &  &  &  \\
\bottomrule
\end{tabular}
}
\end{small}
\end{center}
\vskip -0.1in
\caption{
Failure examples in \compwob{} with advanced models (\texttt{gpt-4}, \texttt{text-davinci-003}).
}
\label{tab:error_gpt4}
\end{table}
\end{landscape}

\section{Per-Task Performance on MiniWoB and CompWoB}
\label{sec:per_task}
\begin{table*}[ht]
\begin{center}
\begin{small}
\scalebox{0.8}{
\begin{tabular}{l|rr}
\toprule
\textbf{Task} & \textbf{HTML-T5}~\citep{gur2023realworld} & \textbf{HTML-T5++} (Ours) \\
\midrule
book-flight & 0.99 & 0.99 \\
choose-date & 0.16 & 1.00 \\
choose-date-easy & 1.00 & 1.00 \\
choose-date-medium & 0.56 & 1.00 \\
choose-list & 0.22 & 1.00 \\
click-button & 1.00 & 1.00 \\
click-button-sequence & 1.00 & 1.00 \\
click-checkboxes & 1.00 & 1.00 \\
click-checkboxes-large & 0.90 & 0.97 \\
click-checkboxes-soft & 0.99 &  1.00\\
click-checkboxes-transfer & 1.00& 1.00 \\
click-collapsible & 1.00 &  1.00\\
click-collapsible-2 & 0.93& 0.96 \\
click-color & 1.00 & 1.00 \\
click-dialog & 1.00 &  1.00\\
click-dialog-2 & 0.74 & 1.00 \\
click-link & 0.99 &  1.00\\
click-menu & 0.37 &  0.96\\
click-option & 1.00 & 1.00 \\
click-pie & 0.96 & 0.94 \\
click-scroll-list & 0.99 & 1.00 \\
click-shades & 0.00 & 1.00 \\
click-shape & 0.79 & 0.95 \\
click-tab & 1.00 & 1.00 \\
click-tab-2 & 0.94 & 0.98 \\
click-tab-2-hard & 0.88 & 0.96 \\
click-test & 1.00 & 1.00 \\
click-test-2 & 1.00 & 1.00 \\
click-widget & 1.00& 1.00 \\
count-shape & 0.67 & 0.92 \\
email-inbox & 1.00 & 0.98 \\
email-inbox-forward-nl & 1.00 & 1.00 \\
email-inbox-forward-nl-turk & 1.00 & 1.00 \\
email-inbox-nl-turk & 0.99 & 1.00 \\
enter-date & 1.00 &  1.00\\
enter-password & 1.00 & 1.00 \\
enter-text & 1.00 & 1.00 \\
enter-text-dynamic & 1.00 & 1.00 \\
enter-time & 1.00 &  1.00\\
focus-text & 1.00 &  1.00\\
focus-text-2 & 1.00 & 1.00 \\
grid-coordinate & 1.00 &1.00  \\
guess-number & 0.13 & 1.00 \\
identify-shape & 1.00 & 1.00 \\
login-user & 1.00 &  1.00\\
login-user-popup & 1.00 & 1.00 \\
multi-layouts & 1.00 &1.00  \\
multi-orderings & 1.00 & 1.00 \\
navigate-tree & 0.99 & 1.00 \\
search-engine & 0.93 & 0.96 \\
social-media & 0.99 &  1.00 \\
social-media-all & 0.31 & 0.31 \\
social-media-some & 0.89 & 0.85 \\
tic-tac-toe & 0.57 & 0.55 \\
use-autocomplete & 0.97& 0.99 \\
use-spinner & 0.07 & 0.06 \\
\midrule
\textbf{Average} & \textbf{0.856} & \textbf{0.952} \\
\bottomrule
\end{tabular}
}
\end{small}
\end{center}
\vskip -0.1in
\caption{Per-task average success rate on 56 tasks from MiniWoB++.
We refer to \citet{gur2023realworld} for the baseline performance.
}
\label{tab:miniwob_per_task}
\end{table*}

\begin{tiny}
\begin{landscape}
\begin{table}[ht]
\begin{center}
\scalebox{0.35}{
\begin{tabular}{l|rrrrrrrrrrrrrrrr}
\toprule
\textbf{Task} & \textbf{WebGUM} & \textbf{HTML-T5} & \textbf{HTML-T5++} & \textbf{RCI} (zero) & \textbf{RCI} (first) & \textbf{RCI} (second) & \textbf{RCI} (comb) & \textbf{RCI} (\texttt{gpt-4}) & \textbf{Synapse} (first) & \textbf{Synapse} (second) & \textbf{Synapse} (best) & \textbf{Synapse} (f;\texttt{gpt-4}) & \textbf{Synapse} (s;\texttt{gpt-4}) & \textbf{Synapse} (b;\texttt{gpt-4}) & \textbf{AdaPlanner} & \textbf{AdaPlanner} (\texttt{davinci})\\
\midrule
click-button\_click-checkboxes & 0.210 & 0.700 & 0.790 & 0.370 & 0.710 & 0.730 & 0.800 & 0.880 & 0.810 & 0.840 & 0.840 & 0.980 & 0.840 & 0.980 & 0.060 & 0.890 \\
click-button\_click-checkboxes-transfer & 0.010 & 0.600 & 0.750 & 0.270 & 0.790 & 0.790 & 0.810 & 1.000 & 0.540 & 0.740 & 0.740 & 0.960 & 0.740 & 0.960 & 0.000 & 0.940 \\
click-button\_click-dialog & 0.870 & 0.720 & 0.970 & 0.740 & 0.830 & 0.820 & 0.940 & 1.000 & 1.000 & 0.650 & 1.000 & 1.000 & 0.660 & 1.000 & 0.900 & 0.980 \\
click-button\_click-link & 0.810 & 0.860 & 0.860 & 0.870 & 0.830 & 0.870 & 0.920 & 0.920 & 0.990 & 0.000 & 0.990 & 1.000 & 0.000 & 1.000 & 0.940 & 0.970 \\
click-button\_click-option & 0.600 & 0.840 & 0.840 & 0.580 & 0.650 & 0.370 & 0.710 & 0.420 & 0.920 & 0.870 & 0.920 & 0.960 & 0.880 & 0.960 & 0.000 & 0.860 \\
click-button-sequence\_click-checkboxes & 0.490 & 0.960 & 0.900 & 0.440 & 0.520 & 0.820 & 0.670 & 0.820 & 0.710 & 0.890 & 0.890 & 0.940 & 0.840 & 0.940 & 0.060 & 0.680 \\
click-button-sequence\_click-option & 0.940 & 1.000 & 1.000 & 0.540 & 0.830 & 0.620 & 0.760 & 0.800 & 0.900 & 0.000 & 0.900 & 0.900 & 0.000 & 0.900 & 0.000 & 0.490 \\
click-button-sequence\_login-user-popup & 0.950 & 0.390 & 0.310 & 0.000 & 0.000 & 0.000 & 0.000 & 0.100 & 0.000 & 0.000 & 0.000 & 0.300 & 0.000 & 0.300 & 0.420 & 0.000 \\
click-link\_click-button & 0.900 & 0.980 & 0.970 & 0.790 & 0.830 & 0.970 & 0.980 & 0.980 & 0.010 & 1.000 & 1.000 & 0.000 & 1.000 & 1.000 & 0.950 & 0.800 \\
click-link\_click-dialog & 0.930 & 0.970 & 1.000 & 0.200 & 0.200 & 0.460 & 0.600 & 0.640 & 0.000 & 0.000 & 0.000 & 0.000 & 0.000 & 0.000 & 0.540 & 0.490 \\
click-link\_click-widget & 0.940 & 0.990 & 0.990 & 0.550 & 0.520 & 0.820 & 0.900 & 0.780 & 0.370 & 0.000 & 0.370 & 0.820 & 0.000 & 0.820 & 0.980 & 0.890 \\
click-link\_enter-text & 0.130 & 0.900 & 1.000 & 0.090 & 0.160 & 0.200 & 0.060 & 0.840 & 0.000 & 0.000 & 0.000 & 0.000 & 0.000 & 0.000 & 0.860 & 0.920 \\
click-option\_enter-text & 1.000 & 0.970 & 1.000 & 0.380 & 0.130 & 0.410 & 0.500 & 0.880 & 0.000 & 0.000 & 0.000 & 1.000 & 0.000 & 1.000 & 0.770 & 0.960 \\
click-option\_login-user & 0.980 & 1.000 & 0.960 & 0.010 & 0.000 & 0.000 & 0.000 & 0.860 & 0.000 & 0.000 & 0.000 & 0.000 & 0.000 & 0.000 & 0.000 & 0.000 \\
click-option\_navigate-tree & 0.020 & 0.480 & 0.370 & 0.550 & 0.470 & 0.500 & 0.720 & 0.820 & 0.000 & 0.240 & 0.240 & 0.560 & 0.960 & 0.960 & 0.620 & 0.590 \\
click-widget\_enter-password & 0.370 & 0.740 & 0.810 & 0.000 & 0.000 & 0.000 & 0.000 & 0.740 & 0.000 & 0.000 & 0.000 & 0.000 & 0.000 & 0.000 & 0.000 & 0.000 \\
click-widget\_multi-layouts & 0.430 & 0.400 & 0.220 & 0.000 & 0.000 & 0.000 & 0.000 & 0.240 & 0.000 & 0.760 & 0.760 & 0.000 & 1.000 & 1.000 & 0.000 & 0.000 \\
enter-password\_click-option & 1.000 & 0.690 & 0.900 & 0.000 & 0.020 & 0.000 & 0.000 & 0.780 & 0.000 & 0.000 & 0.000 & 0.000 & 0.000 & 0.000 & 0.000 & 0.000 \\
login-user\_navigate-tree & 0.000 & 0.040 & 0.140 & 0.020 & 0.010 & 0.000 & 0.000 & 0.700 & 0.000 & 0.000 & 0.000 & 0.000 & 0.000 & 0.000 & 0.000 & 0.000 \\
multi-layouts\_login-user & 0.000 & 0.000 & 0.000 & 0.000 & 0.000 & 0.000 & 0.000 & 0.100 & 0.720 & 0.000 & 0.720 & 1.000 & 0.000 & 1.000 & 0.010 & 0.000 \\
\midrule
\textbf{Average (two-way)} & \textbf{0.579} & \textbf{0.712} & \textbf{0.739} & \textbf{0.320} & \textbf{0.375} & \textbf{0.419} & \textbf{0.469} & \textbf{0.715} & \textbf{0.349} & \textbf{0.300} & \textbf{0.469} & \textbf{0.521} & \textbf{0.346} & \textbf{0.641} & \textbf{0.356} & \textbf{0.523} \\
\midrule
click-button\_click-option\_login-user & 0.140 & 0.520 & 0.740 & 0.000 & 0.340 & 0.000 & 0.000 & 0.120 & 0.000 & 0.000 & 0.000 & 0.940 & 0.280 & 0.940 & 0.140 & 0.000 \\
click-button-sequence\_click-option\_login-user & 0.300 & 0.200 & 0.770 & 0.520 & 0.740 & 0.680 & 0.840 & 0.560 & 0.720 & 0.000 & 0.720 & 0.660 & 0.000 & 0.660 & 0.000 & 0.000 \\
click-checkboxes\_click-widget\_click-button-sequence & 0.200 & 0.580 & 0.780 & 0.400 & 0.660 & 0.660 & 0.820 & 1.000 & 0.820 & 0.000 & 0.820 & 0.820 & 0.180 & 0.820 & 0.580 & 0.720 \\
click-checkboxes-transfer\_click-button-sequence\_enter-password & 0.000 & 0.010 & 0.010 & 0.000 & 0.000 & 0.000 & 0.000 & 0.660 & 0.000 & 0.000 & 0.000 & 0.000 & 0.500 & 0.500 & 0.000 & 0.000 \\
click-checkboxes-transfer\_enter-password\_click-dialog & 0.000 & 0.010 & 0.000 & 0.000 & 0.000 & 0.000 & 0.000 & 0.280 & 0.000 & 0.420 & 0.420 & 0.000 & 0.400 & 0.400 & 0.000 & 0.000 \\
click-dialog\_click-button-sequence\_enter-password & 0.890 & 1.000 & 1.000 & 0.000 & 0.000 & 0.000 & 0.000 & 0.700 & 0.000 & 0.000 & 0.000 & 0.000 & 0.000 & 0.000 & 0.000 & 0.000 \\
click-dialog\_click-checkboxes-transfer\_click-widget & 0.140 & 0.690 & 0.560 & 0.140 & 0.540 & 0.340 & 0.400 & 0.680 & 0.000 & 0.000 & 0.000 & 0.000 & 0.000 & 0.000 & 0.000 & 0.000 \\
click-link\_click-button\_click-dialog & 0.730 & 0.940 & 0.980 & 0.240 & 0.270 & 0.480 & 0.610 & 0.600 & 0.000 & 0.600 & 0.600 & 0.000 & 0.620 & 0.620 & 0.560 & 0.580 \\
click-widget\_click-option\_click-dialog & 0.010 & 0.120 & 0.410 & 0.120 & 0.000 & 0.020 & 0.300 & 0.400 & 0.000 & 0.000 & 0.000 & 0.020 & 0.000 & 0.020 & 0.000 & 0.000 \\
enter-password\_click-checkboxes\_login-user-popup & 0.110 & 0.000 & 0.060 & 0.000 & 0.000 & 0.000 & 0.000 & 0.700 & 0.000 & 0.000 & 0.000 & 0.000 & 0.000 & 0.000 & 0.000 & 0.000 \\
\midrule
\textbf{Average (three-way)} & \textbf{0.252} & \textbf{0.407} & \textbf{0.531} & \textbf{0.142} & \textbf{0.255} & \textbf{0.218} & \textbf{0.297} & \textbf{0.570} & \textbf{0.154} & \textbf{0.102} & \textbf{0.256} & \textbf{0.244} & \textbf{0.198} & \textbf{0.396} & \textbf{0.128} & \textbf{0.130} \\
\midrule
click-button-sequence\_click-widget\_click-link\_click-button\_click-checkboxes\_click-option\_click-dialog & 0.000 & 0.230 & 0.320 & 0.160 & 0.020 & 0.380 & 0.080 & 0.720 & 0.000 & 0.000 & 0.000 & 0.000 & 0.000 & 0.000 & 0.000 & 0.000 \\
click-button-sequence\_click-widget\_click-link\_click-button\_click-checkboxes\_click-option\_click-dialog\_login-user & 0.000 & 0.000 & 0.000 & 0.000 & 0.000 & 0.000 & 0.000 & 0.620 & 0.000 & 0.000 & 0.000 & 0.000 & 0.000 & 0.000 & 0.000 & 0.000 \\
click-link\_click-button\_click-checkboxes\_click-dialog & 0.280 & 0.630 & 0.810 & 0.150 & 0.230 & 0.320 & 0.370 & 0.540 & 0.000 & 0.310 & 0.310 & 0.000 & 0.600 & 0.600 & 0.050 & 0.600 \\
click-link\_click-button\_click-checkboxes\_click-option\_click-dialog & 0.180 & 0.430 & 0.780 & 0.200 & 0.180 & 0.370 & 0.360 & 0.560 & 0.000 & 0.060 & 0.060 & 0.000 & 0.620 & 0.620 & 0.020 & 0.490 \\
click-widget\_click-link\_click-button\_click-checkboxes\_click-option\_click-dialog & 0.000 & 0.040 & 0.140 & 0.140 & 0.360 & 0.100 & 0.300 & 0.480 & 0.000 & 0.000 & 0.000 & 0.000 & 0.000 & 0.000 & 0.000 & 0.000 \\
\midrule
\textbf{Average (n-way)} & \textbf{0.092} & \textbf{0.266} & \textbf{0.410} & \textbf{0.130} & \textbf{0.158} & \textbf{0.234} & \textbf{0.222} & \textbf{0.584} & \textbf{0.000} & \textbf{0.074} & \textbf{0.074} & \textbf{0.000} & \textbf{0.244} & \textbf{0.244} & \textbf{0.014} & \textbf{0.218} \\
\midrule
click-checkboxes-transfer\_multi-layouts\_email-inbox-forward-nl-transition & 0.240 & 0.460 & 0.650 & 0.000 & 0.000 & 0.000 & 0.000 & 0.000 & 0.000 & 0.000 & 0.000 & 0.000 & 0.000 & 0.000 & 0.220 & 0.440 \\
click-option\_login-user-transition & 0.990 & 1.000 & 0.800 & 0.000 & 0.000 & 0.000 & 0.000 & 0.000 & 0.000 & 0.000 & 0.000 & 0.000 & 0.000 & 0.000 & 0.000 & 0.000 \\
click-option\_multi-layouts\_click-widget\_login-user-popup-transition & 0.190 & 0.220 & 0.130 & 0.000 & 0.000 & 0.000 & 0.000 & 0.000 & 0.000 & 0.000 & 0.000 & 0.000 & 0.000 & 0.000 & 0.000 & 0.000 \\
login-user\_navigate-tree-transition & 0.960 & 1.000 & 1.000 & 0.000 & 0.000 & 0.000 & 0.000 & 0.460 & 0.000 & 0.000 & 0.000 & 0.340 & 0.000 & 0.340 & 0.000 & 0.000 \\
login-user-popup\_email-inbox-forward-nl-turk-transition & 1.000 & 0.960 & 0.910 & 0.000 & 0.000 & 0.200 & 0.000 & 0.120 & 0.000 & 0.000 & 0.000 & 0.000 & 0.000 & 0.000 & 0.700 & 0.560 \\
\midrule
\textbf{Average (transition)} & \textbf{0.676} & \textbf{0.728} & \textbf{0.698} & \textbf{0.000} & \textbf{0.000} & \textbf{0.040} & \textbf{0.000} & \textbf{0.116} & \textbf{0.000} & \textbf{0.000} & \textbf{0.000} & \textbf{0.068} & \textbf{0.000} & \textbf{0.068} & \textbf{0.184} & \textbf{0.200} \\
\midrule
click-button\_click-tab-2-hard & 0.240 & 0.550 & 0.510 & 0.080 & 0.120 & 0.260 & 0.340 & 0.580 & 0.280 & 0.040 & 0.280 & 0.960 & 0.320 & 0.960 & 0.260 & 0.280 \\
click-button-sequence\_use-autocomplete & 0.410 & 0.220 & 0.840 & 0.200 & 0.060 & 0.640 & 0.160 & 0.820 & 0.000 & 0.000 & 0.000 & 0.420 & 0.000 & 0.420 & 0.000 & 0.000 \\
click-checkboxes-soft\_enter-password & 0.990 & 0.850 & 0.910 & 0.000 & 0.000 & 0.000 & 0.000 & 0.820 & 0.000 & 0.000 & 0.000 & 0.000 & 0.000 & 0.000 & 0.000 & 0.000 \\
click-checkboxes-soft\_multi-layouts & 0.930 & 0.240 & 0.170 & 0.000 & 0.000 & 0.000 & 0.000 & 0.120 & 0.000 & 0.300 & 0.000 & 0.000 & 0.820 & 0.820 & 0.000 & 0.000 \\
click-dialog\_search-engine & 0.930 & 0.970 & 0.940 & 0.100 & 0.020 & 0.000 & 0.000 & 0.920 & 0.000 & 0.000 & 0.000 & 0.000 & 0.000 & 0.000 & 0.000 & 0.000 \\
click-dialog-2\_click-widget & 0.270 & 0.460 & 0.690 & 0.000 & 0.000 & 0.000 & 0.040 & 0.020 & 0.100 & 0.000 & 0.100 & 0.180 & 0.000 & 0.180 & 0.160 & 0.100 \\
click-dialog-2\_login-user-popup & 0.330 & 0.440 & 0.360 & 0.000 & 0.000 & 0.000 & 0.000 & 0.060 & 0.000 & 0.000 & 0.000 & 0.000 & 0.000 & 0.000 & 0.000 & 0.000 \\
click-widget\_click-checkboxes-soft & 0.120 & 0.290 & 0.460 & 0.080 & 0.100 & 0.060 & 0.200 & 0.600 & 0.000 & 0.000 & 0.000 & 0.000 & 0.000 & 0.000 & 0.100 & 0.140 \\
enter-date\_login-user & 0.990 & 1.000 & 0.230 & 0.000 & 0.020 & 0.000 & 0.060 & 0.500 & 0.000 & 0.000 & 0.000 & 0.000 & 0.000 & 0.000 & 0.380 & 0.380 \\
use-autocomplete\_click-dialog & 0.000 & 0.000 & 0.000 & 0.000 & 0.080 & 0.000 & 0.080 & 0.060 & 0.000 & 0.000 & 0.000 & 0.000 & 0.000 & 0.000 & 0.000 & 0.000 \\
\midrule
\textbf{Average (two-way, easy-medium)} & \textbf{0.521} & \textbf{0.502} & \textbf{0.511} & \textbf{0.046} & \textbf{0.040} & \textbf{0.096} & \textbf{0.088} & \textbf{0.450} & \textbf{0.038} & \textbf{0.034} & \textbf{0.038} & \textbf{0.156} & \textbf{0.114} & \textbf{0.238} & \textbf{0.090} & \textbf{0.090} \\
\textbf{Average (total)} & \textbf{0.463} & \textbf{0.566}& \textbf{0.615} & \textbf{0.179} & \textbf{0.225} & \textbf{0.258} & \textbf{0.287} & \textbf{0.560} & \textbf{0.178} & \textbf{0.154} & \textbf{0.254} & \textbf{0.295} & \textbf{0.225} & \textbf{0.414} & \textbf{0.206} & \textbf{0.295} \\
\bottomrule
\end{tabular}
}
\end{center}
\caption{
Per-task success rate on \compwob{}.
}
\label{tab:per_task_compwob_results}
\end{table}
\end{landscape}
\end{tiny}

\begin{tiny}
\begin{landscape}
\begin{table}[ht]
\begin{center}
\scalebox{0.45}{
\begin{tabular}{l|rrrrrrrrrrrr}
\toprule
\textbf{Task} &\textbf{WebGUM} & \textbf{HTML-T5} & \textbf{HTML-T5++} & \textbf{RCI} (zero) & \textbf{RCI} (first) & \textbf{RCI} (second) & \textbf{RCI} (comb) & \textbf{RCI} (\texttt{gpt-4}) & \textbf{Synapse} (first) & \textbf{Synapse} (second) & \textbf{Synapse} (best) & \textbf{AdaPlanner} \\
\midrule
click-button\_click-checkboxes & 0.180 & 0.340 & 0.630 & 0.420 & 0.310 & 0.420 & 0.380 & 0.520 & 0.670 & 0.800 & 0.800 & 0.340\\
click-button\_click-checkboxes-transfer & 0.060 & 0.260 & 0.630 & 0.150 & 0.140 & 0.230 & 0.200 & 0.320 & 0.360 & 0.930 & 0.930 & 0.090\\
click-button\_click-dialog & 0.300 & 0.020 & 0.030 & 0.850 & 0.790 & 0.790 & 0.810 & 0.840 & 0.170 & 0.200 & 0.200 & 0.000\\
click-button\_click-link & 0.010 & 0.050 & 0.040 & 0.430 & 0.500 & 0.430 & 0.460 & 0.560 & 0.870 & 0.000 & 0.870 & 0.970\\
click-button\_click-option & 0.230 & 0.420 & 0.550 & 0.350 & 0.270 & 0.400 & 0.410 & 0.660 & 0.790 & 0.860 & 0.860 & 0.440\\
click-button-sequence\_click-checkboxes & 0.000 & 0.140 & 0.130 & 0.270 & 0.260 & 0.390 & 0.280 & 0.820 & 0.480 & 0.870 & 0.870 & 0.700\\
click-button-sequence\_click-option & 0.000 & 0.020 & 0.110 & 0.360 & 0.420 & 0.320 & 0.490 & 0.920 & 0.350 & 0.000 & 0.350 & 0.550\\
click-button-sequence\_login-user-popup & 0.000 & 0.720 & 0.020 & 0.000 & 0.000 & 0.000 & 0.000 & 0.160 & 0.000 & 0.000 & 0.000 & 0.280\\
click-link\_click-button & 0.910 & 0.750 & 0.580 & 0.840 & 0.760 & 0.800 & 0.830 & 0.920 & 0.000 & 0.670 & 0.670 & 0.950\\
click-link\_click-dialog & 0.450 & 0.050 & 0.230 & 0.330 & 0.430 & 0.470 & 0.880 & 0.660 & 0.000 & 0.000 & 0.000 & 0.200\\
click-link\_click-widget & 0.140 & 0.750 & 0.740 & 0.550 & 0.580 & 0.480 & 0.530 & 0.780 & 0.640 & 0.000 & 0.640 & 0.920\\
click-link\_enter-text & 0.000 & 0.950 & 0.910 & 0.240 & 0.280 & 0.300 & 0.310 & 0.480 & 0.000 & 0.000 & 0.000 & 0.970\\
click-option\_enter-text & 1.000 & 0.660 & 0.750 & 0.570 & 0.650 & 0.650 & 0.640 & 0.800 & 0.000 & 0.000 & 0.000 & 0.590\\
click-option\_login-user & 0.980 & 0.220 & 0.150 & 0.010 & 0.000 & 0.010 & 0.000 & 0.620 & 0.000 & 0.000 & 0.000 & 0.020\\
click-option\_navigate-tree & 0.000 & 0.060 & 0.050 & 0.450 & 0.390 & 0.500 & 0.490 & 0.440 & 0.050 & 0.590 & 0.590 & 0.880\\
click-widget\_enter-password & 0.280 & 0.470 & 0.530 & 0.000 & 0.000 & 0.000 & 0.000 & 0.300 & 0.000 & 0.000 & 0.000 & 0.000\\
click-widget\_multi-layouts & 0.260 & 0.310 & 0.220 & 0.000 & 0.000 & 0.000 & 0.000 & 0.440 & 0.000 & 0.700 & 0.700 & 0.020\\
enter-password\_click-option & 0.960 & 0.160 & 0.900 & 0.000 & 0.000 & 0.000 & 0.000 & 0.920 & 0.000 & 0.000 & 0.000 & 0.000\\
login-user\_navigate-tree & 0.000 & 0.030 & 0.080 & 0.010 & 0.000 & 0.000 & 0.000 & 0.280 & 0.000 & 0.000 & 0.000 & 0.000\\
multi-layouts\_login-user & 0.000 & 0.000 & 0.000 & 0.000 & 0.000 & 0.000 & 0.000 & 0.020 & 0.210 & 0.000 & 0.210 & 0.000\\
\midrule
\textbf{Average (two-way)} & \textbf{0.288} & \textbf{0.319} & \textbf{0.364} & \textbf{0.292} & \textbf{0.289} & \textbf{0.310} & \textbf{0.336} & \textbf{0.573} & \textbf{0.230} & \textbf{0.281} & \textbf{0.385} & \textbf{0.396}\\
\midrule
click-button\_click-option\_login-user & 0.250 & 0.140 & 0.220 & 0.020 & 0.000 & 0.000 & 0.000 & 0.200 & 0.000 & 0.000 & 0.000 & 0.220\\
click-button-sequence\_click-option\_login-user & 0.000 & 0.000 & 0.220 & 0.420 & 0.800 & 0.440 & 0.740 & 0.000 & 0.000 & 0.000 & 0.000 & 0.320\\
click-checkboxes\_click-widget\_click-button-sequence & 0.000 & 0.270 & 0.100 & 0.060 & 0.140 & 0.140 & 0.080 & 0.820 & 0.000 & 0.000 & 0.000 & 0.000\\
click-checkboxes-transfer\_click-button-sequence\_enter-password & 0.000 & 0.000 & 0.000 & 0.000 & 0.000 & 0.000 & 0.000 & 0.340 & 0.000 & 0.000 & 0.000 & 0.000\\
click-checkboxes-transfer\_enter-password\_click-dialog & 0.000 & 0.000 & 0.000 & 0.000 & 0.000 & 0.000 & 0.000 & 0.000 & 0.000 & 0.000 & 0.000 & 0.000\\
click-dialog\_click-button-sequence\_enter-password & 1.000 & 0.690 & 0.040 & 0.000 & 0.000 & 0.000 & 0.000 & 0.680 & 0.000 & 0.000 & 0.000 & 0.000\\
click-dialog\_click-checkboxes-transfer\_click-widget & 0.080 & 0.520 & 0.460 & 0.140 & 0.340 & 0.000 & 0.360 & 0.580 & 0.000 & 0.000 & 0.000 & 0.000\\
click-link\_click-button\_click-dialog & 0.040 & 0.050 & 0.130 & 0.180 & 0.320 & 0.320 & 0.310 & 0.360 & 0.000 & 0.000 & 0.000 & 0.000\\
click-widget\_click-option\_click-dialog & 0.000 & 0.000 & 0.000 & 0.120 & 0.220 & 0.160 & 0.140 & 0.120 & 0.000 & 0.000 & 0.000 & 0.000\\
enter-password\_click-checkboxes\_login-user-popup & 0.000 & 0.000 & 0.000 & 0.000 & 0.000 & 0.000 & 0.000 & 0.660 & 0.000 & 0.000 & 0.000 & 0.000\\
\midrule
\textbf{Average (three-way)} & \textbf{0.137} & \textbf{0.167} & \textbf{0.117} & \textbf{0.094} & \textbf{0.182} & \textbf{0.106} & \textbf{0.163} & \textbf{0.376} & \textbf{0.000} & \textbf{0.000} & \textbf{0.000} & \textbf{0.054} \\
\midrule
click-button-sequence\_click-widget\_click-link\_click-button\_click-checkboxes\_click-option\_click-dialog & 0.000 & 0.070 & 0.090 & 0.140 & 0.100 & 0.340 & 0.080 & 0.600 & 0.000 & 0.000 & 0.000 & 0.000\\
click-button-sequence\_click-widget\_click-link\_click-button\_click-checkboxes\_click-option\_click-dialog\_login-user & 0.000 & 0.000 & 0.000 & 0.000 & 0.000 & 0.000 & 0.000 & 0.180 & 0.000 & 0.000 & 0.000 & 0.000\\
click-link\_click-button\_click-checkboxes\_click-dialog & 0.000 & 0.090 & 0.240 & 0.220 & 0.220 & 0.120 & 0.200 & 0.320 & 0.000 & 0.000 & 0.000 & 0.000\\
click-link\_click-button\_click-checkboxes\_click-option\_click-dialog & 0.000 & 0.060 & 0.260 & 0.170 & 0.120 & 0.140 & 0.170 & 0.380 & 0.000 & 0.000 & 0.000 & 0.000\\
click-widget\_click-link\_click-button\_click-checkboxes\_click-option\_click-dialog & 0.000 & 0.000 & 0.030 & 0.060 & 0.060 & 0.040 & 0.180 & 0.080 & 0.000 & 0.000 & 0.000 & 0.000\\
\midrule
\textbf{Average (n-way)} & \textbf{0.000} & \textbf{0.044} & \textbf{0.124} & \textbf{0.118} & \textbf{0.100} & \textbf{0.128} & \textbf{0.126} & \textbf{0.312} & \textbf{0.000} & \textbf{0.000} & \textbf{0.000} & \textbf{0.000} \\
\midrule
click-checkboxes-transfer\_multi-layouts\_email-inbox-forward-nl-transition & 0.170 & 0.360 & 0.780 & 0.000 & 0.000 & 0.000 & 0.000 & 0.000 & 0.000 & 0.000 & 0.000 & 0.140\\
click-option\_login-user-transition & 0.990 & 0.800 & 0.650 & 0.000 & 0.000 & 0.000 & 0.000 & 0.080 & 0.000 & 0.000 & 0.000 & 0.000\\
click-option\_multi-layouts\_click-widget\_login-user-popup-transition & 0.070 & 0.100 & 0.030 & 0.000 & 0.000 & 0.000 & 0.000 & 0.000 & 0.000 & 0.000 & 0.000 & 0.000\\
login-user\_navigate-tree-transition & 1.000 & 1.000 & 0.990 & 0.000 & 0.000 & 0.000 & 0.000 & 0.080 & 0.000 & 0.000 & 0.000 & 0.000\\
login-user-popup\_email-inbox-forward-nl-turk-transition & 0.650 & 0.770 & 0.820 & 0.000 & 0.000 & 0.220 & 0.100 & 0.300 & 0.000 & 0.000 & 0.000 & 0.580\\
\midrule
\textbf{Average (transition)} & \textbf{0.576} & \textbf{0.606} & \textbf{0.654} & \textbf{0.000} & \textbf{0.000} & \textbf{0.044} & \textbf{0.020} & \textbf{0.092} & \textbf{0.000} & \textbf{0.000} & \textbf{0.000} & \textbf{0.144} \\
\midrule
click-button\_click-tab-2-hard & 0.050 & 0.470 & 0.490 & 0.140 & 0.120 & 0.180 & 0.400 & 0.640 & 0.300 & 0.020 & 0.300 & 0.160\\
click-button-sequence\_use-autocomplete & 0.000 & 0.080 & 0.930 & 0.260 & 0.080 & 0.640 & 0.040 & 0.860 & 0.000 & 0.000 & 0.000 & 0.000\\
click-checkboxes-soft\_enter-password & 0.950 & 0.490 & 0.760 & 0.000 & 0.000 & 0.000 & 0.000 & 0.740 & 0.000 & 0.000 & 0.000 & 0.000\\
click-checkboxes-soft\_multi-layouts & 0.940 & 0.390 & 0.260 & 0.000 & 0.000 & 0.000 & 0.000 & 0.200 & 0.000 & 0.420 & 0.000 & 0.000\\
click-dialog\_search-engine & 0.950 & 0.820 & 0.550 & 0.000 & 0.000 & 0.000 & 0.020 & 0.980 & 0.000 & 0.000 & 0.000 & 0.000\\
click-dialog-2\_click-widget & 0.210 & 0.370 & 0.260 & 0.000 & 0.000 & 0.020 & 0.000 & 0.020 & 0.000 & 0.000 & 0.000 & 0.080\\
click-dialog-2\_login-user-popup & 0.380 & 0.360 & 0.330 & 0.000 & 0.000 & 0.000 & 0.000 & 0.020 & 0.000 & 0.000 & 0.000 & 0.000\\
click-widget\_click-checkboxes-soft & 0.090 & 0.240 & 0.290 & 0.080 & 0.080 & 0.120 & 0.080 & 0.520 & 0.000 & 0.000 & 0.000 & 0.000\\
enter-date\_login-user & 0.950 & 0.350 & 0.960 & 0.000 & 0.040 & 0.000 & 0.000 & 0.400 & 0.000 & 0.000 & 0.000 & 0.040\\
use-autocomplete\_click-dialog & 0.000 & 0.000 & 0.000 & 0.000 & 0.080 & 0.000 & 0.000 & 0.120 & 0.000 & 0.000 & 0.000 & 0.000\\
\midrule
\textbf{Average (two-way, easy-medium)} & \textbf{0.452} & \textbf{0.357} & \textbf{0.483} & \textbf{0.048} & \textbf{0.040} & \textbf{0.096} & \textbf{0.054} & \textbf{0.450} & \textbf{0.030} & \textbf{0.044} & \textbf{0.030} & \textbf{0.028}\\
\textbf{Average (total)} & \textbf{0.291} & \textbf{0.297} & \textbf{0.343} & \textbf{0.157} & \textbf{0.170} & \textbf{0.181} & \textbf{0.192} & \textbf{0.435} & \textbf{0.098} & \textbf{0.121} & \textbf{0.160} & \textbf{0.189} \\
\bottomrule
\end{tabular}
}
\end{center}
\caption{
Per-task success rate on \compwob{} with reverse-order instructions.
}
\label{tab:per_task_reverse_results}
\end{table}
\end{landscape}
\end{tiny}

\begin{tiny}
\begin{landscape}
\begin{table}[ht]
\begin{center}
\scalebox{0.34}{
\begin{tabular}{l|rrr|r|rrrrrrrrrrrrr}
\toprule
\textbf{Task} & \textbf{RCI} (optimal) & \textbf{Synapse} (optimal) & \textbf{AdaPlanner} (optimal) & \textbf{RCI} (oracle, \texttt{gpt-4}) & \textbf{RCI} (zero) & \textbf{RCI} (first) & \textbf{RCI} (second) & \textbf{RCI} (comb) & \textbf{RCI} (\texttt{gpt-4}) & \textbf{Synapse} (first) & \textbf{Synapse} (second) & \textbf{Synapse} (best) & \textbf{Synapse} (f;\texttt{gpt-4}) & \textbf{Synapse} (s;\texttt{gpt-4}) & \textbf{Synapse} (b;\texttt{gpt-4}) & \textbf{AdaPlanner} & \textbf{AdaPlanner} (\texttt{davinci})\\
\midrule
click-button\_click-checkboxes & 1.000 & 1.000 & 1.000 & 1.000 & 0.370 & 0.710 & 0.730 & 0.800 & 0.880 & 0.810 & 0.840 & 0.840 & 0.980 & 0.840 & 0.980 & 0.060 & 0.890 \\
click-button\_click-checkboxes-transfer & 1.000 & 1.000 & 1.000 & 0.980 & 0.270 & 0.790 & 0.790 & 0.810 & 1.000 & 0.540 & 0.740 & 0.740 & 0.960 & 0.740 & 0.960 & 0.000 & 0.940 \\
click-button\_click-dialog & 1.000 & 1.000 & 1.000 & 1.000 & 0.740 & 0.830 & 0.820 & 0.940 & 1.000 & 1.000 & 0.650 & 1.000 & 1.000 & 0.660 & 1.000 & 0.900 & 0.980 \\
click-button\_click-link & 1.000 & 1.00 & 0.980 & 0.960 & 0.870 & 0.830 & 0.870 & 0.920 & 0.920 & 0.990 & 0.000 & 0.990 & 1.000 & 0.000 & 1.000 & 0.940 & 0.970 \\
click-button\_click-option & 1.000 & 1.000 & 1.000 & 1.000 & 0.580 & 0.650 & 0.370 & 0.710 & 0.420 & 0.920 & 0.870 & 0.920 & 0.960 & 0.880 & 0.960 & 0.000 & 0.860 \\
click-button-sequence\_click-checkboxes & 1.000 & 1.000 & 1.000 & 0.980 & 0.440 & 0.520 & 0.820 & 0.670 & 0.820 & 0.710 & 0.890 & 0.890 & 0.940 & 0.840 & 0.940 & 0.060 & 0.680 \\
click-button-sequence\_click-option & 1.000 & 1.000 & 1.000 & 1.000 & 0.540 & 0.830 & 0.620 & 0.760 & 0.800 & 0.900 & 0.000 & 0.900 & 0.900 & 0.000 & 0.900 & 0.000 & 0.490 \\
click-button-sequence\_login-user-popup & 1.000 & 1.000 & 1.000 & 1.000 & 0.000 & 0.000 & 0.000 & 0.000 & 0.100 & 0.000 & 0.000 & 0.000 & 0.300 & 0.000 & 0.300 & 0.420 & 0.000 \\
click-link\_click-button & 1.000 & 1.000 & 0.980 & 0.980 & 0.790 & 0.830 & 0.970 & 0.980 & 0.980 & 0.010 & 1.000 & 1.000 & 0.000 & 1.000 & 1.000 & 0.950 & 0.800 \\
click-link\_click-dialog & 1.000 & 1.000 & 0.980 & 0.660 & 0.200 & 0.200 & 0.460 & 0.600 & 0.640 & 0.000 & 0.000 & 0.000 & 0.000 & 0.000 & 0.000 & 0.540 & 0.490 \\
click-link\_click-widget & 0.980 & 1.000 & 0.980 & 1.000 & 0.550 & 0.520 & 0.820 & 0.900 & 0.780 & 0.370 & 0.000 & 0.370 & 0.820 & 0.000 & 0.820 & 0.980 & 0.890 \\
click-link\_enter-text & 1.000 & 1.000 & 0.960 & 1.000 & 0.090 & 0.160 & 0.200 & 0.060 & 0.840 & 0.000 & 0.000 & 0.000 & 0.000 & 0.000 & 0.000 & 0.860 & 0.920 \\
click-option\_enter-text & 1.000 & 1.000 & 0.980 & 1.000 & 0.380 & 0.130 & 0.410 & 0.500 & 0.880 & 0.000 & 0.000 & 0.000 & 1.000 & 0.000 & 1.000 & 0.770 & 0.960 \\
click-option\_login-user & 1.000 & 1.000 & 1.000 & 1.000 & 0.010 & 0.000 & 0.000 & 0.000 & 0.860 & 0.000 & 0.000 & 0.000 & 0.000 & 0.000 & 0.000 & 0.000 & 0.000 \\
click-option\_navigate-tree & 0.860 & 0.960 & 0.820 & 0.960 & 0.550 & 0.470 & 0.500 & 0.720 & 0.820 & 0.000 & 0.240 & 0.240 & 0.560 & 0.960 & 0.960 & 0.620 & 0.590 \\
click-widget\_enter-password & 0.980 & 1.000 & 0.980 & 0.760 & 0.000 & 0.000 & 0.000 & 0.000 & 0.740 & 0.000 & 0.000 & 0.000 & 0.000 & 0.000 & 0.000 & 0.000 & 0.000 \\
click-widget\_multi-layouts & 0.706 & 0.940 & 0.840 & 0.400 & 0.000 & 0.000 & 0.000 & 0.000 & 0.240 & 0.000 & 0.760 & 0.760 & 0.000 & 1.000 & 1.000 & 0.000 & 0.000 \\
enter-password\_click-option & 1.000 & 1.000 & 0.980 & 0.980 & 0.000 & 0.020 & 0.000 & 0.000 & 0.780 & 0.000 & 0.000 & 0.000 & 0.000 & 0.000 & 0.000 & 0.000 & 0.000 \\
login-user\_navigate-tree & 0.860 & 0.960 & 0.820 & 0.780 & 0.020 & 0.010 & 0.000 & 0.000 & 0.700 & 0.000 & 0.000 & 0.000 & 0.000 & 0.000 & 0.000 & 0.000 & 0.000 \\
multi-layouts\_login-user & 0.720 & 0.940 & 0.840 & 0.100 & 0.000 & 0.000 & 0.000 & 0.000 & 0.100 & 0.720 & 0.000 & 0.720 & 1.000 & 0.000 & 1.000 & 0.010 & 0.000 \\
\midrule
\textbf{Average (two-way)} & \textbf{0.939} & \textbf{0.990} & \textbf{0.955} & \textbf{0.833} & \textbf{0.320} & \textbf{0.375} & \textbf{0.419} & \textbf{0.469} & \textbf{0.715} & \textbf{0.349} & \textbf{0.300} & \textbf{0.469} & \textbf{0.521} & \textbf{0.346} & \textbf{0.641} & \textbf{0.356} & \textbf{0.523} \\
\midrule
click-button\_click-option\_login-user & 1.000 & 1.000 & 1.000 & -- & 0.000 & 0.340 & 0.000 & 0.000 & 0.120 & 0.000 & 0.000 & 0.000 & 0.940 & 0.280 & 0.940 & 0.140 & 0.000 \\
click-button-sequence\_click-option\_login-user & 1.000 & 1.000 & 1.000 & -- & 0.520 & 0.740 & 0.680 & 0.840 & 0.560 & 0.720 & 0.000 & 0.720 & 0.660 & 0.000 & 0.660 & 0.000 & 0.000 \\
click-checkboxes\_click-widget\_click-button-sequence & 0.980 & 1.000 & 1.000 & -- & 0.400 & 0.660 & 0.660 & 0.820 & 1.000 & 0.820 & 0.000 & 0.820 & 0.820 & 0.180 & 0.820 & 0.580 & 0.720 \\
click-checkboxes-transfer\_click-button-sequence\_enter-password & 1.000 & 1.000 & 0.960 & -- & 0.000 & 0.000 & 0.000 & 0.000 & 0.660 & 0.000 & 0.000 & 0.000 & 0.000 & 0.500 & 0.500 & 0.000 & 0.000 \\
click-checkboxes-transfer\_enter-password\_click-dialog & 1.000 & 1.000 & 0.960 & -- & 0.000 & 0.000 & 0.000 & 0.000 & 0.280 & 0.000 & 0.420 & 0.420 & 0.000 & 0.400 & 0.400 & 0.000 & 0.000 \\
click-dialog\_click-button-sequence\_enter-password & 1.000 & 1.000 & 0.980 & -- & 0.000 & 0.000 & 0.000 & 0.000 & 0.700 & 0.000 & 0.000 & 0.000 & 0.000 & 0.000 & 0.000 & 0.000 & 0.000 \\
click-dialog\_click-checkboxes-transfer\_click-widget & 0.980 & 1.000 & 0.980 & -- & 0.140 & 0.540 & 0.340 & 0.400 & 0.680 & 0.000 & 0.000 & 0.000 & 0.000 & 0.000 & 0.000 & 0.000 & 0.000 \\
click-link\_click-button\_click-dialog & 1.000 & 1.000 & 0.980 & -- & 0.240 & 0.270 & 0.480 & 0.610 & 0.600 & 0.000 & 0.600 & 0.600 & 0.000 & 0.620 & 0.620 & 0.560 & 0.580 \\
click-widget\_click-option\_click-dialog & 0.980 & 1.000 & 1.000 & -- & 0.120 & 0.000 & 0.020 & 0.300 & 0.400 & 0.000 & 0.000 & 0.000 & 0.020 & 0.000 & 0.020 & 0.000 & 0.000 \\
enter-password\_click-checkboxes\_login-user-popup & 0.680 & 1.000 & 0.960 & -- & 0.000 & 0.000 & 0.000 & 0.000 & 0.700 & 0.000 & 0.000 & 0.000 & 0.000 & 0.000 & 0.000 & 0.000 & 0.000 \\
\midrule
\textbf{Average (three-way)} & \textbf{0.962} & \textbf{1.000} & \textbf{0.982} & -- & \textbf{0.142} & \textbf{0.255} & \textbf{0.218} & \textbf{0.297} & \textbf{0.570} & \textbf{0.154} & \textbf{0.102} & \textbf{0.256} & \textbf{0.244} & \textbf{0.198} & \textbf{0.396} & \textbf{0.128} & \textbf{0.130} \\
\midrule
click-button-sequence\_click-widget\_click-link\_click-button\_click-checkboxes\_click-option\_click-dialog & 0.980 & 1.000 & 0.980 & -- & 0.160 & 0.020 & 0.380 & 0.080 & 0.720 & 0.000 & 0.000 & 0.000 & 0.000 & 0.000 & 0.000 & 0.000 & 0.000 \\
click-button-sequence\_click-widget\_click-link\_click-button\_click-checkboxes\_click-option\_click-dialog\_login-user & 0.980 & 1.000 & 0.980 & -- & 0.000 & 0.000 & 0.000 & 0.000 & 0.620 & 0.000 & 0.000 & 0.000 & 0.000 & 0.000 & 0.000 & 0.000 & 0.000 \\
click-link\_click-button\_click-checkboxes\_click-dialog & 1.000 & 1.000 & 0.980 & -- & 0.150 & 0.230 & 0.320 & 0.370 & 0.540 & 0.000 & 0.310 & 0.310 & 0.000 & 0.600 & 0.600 & 0.050 & 0.600 \\
click-link\_click-button\_click-checkboxes\_click-option\_click-dialog & 1.000 & 1.000 & 0.980 & -- & 0.200 & 0.180 & 0.370 & 0.360 & 0.560 & 0.000 & 0.060 & 0.060 & 0.000 & 0.620 & 0.620 & 0.020 & 0.490 \\
click-widget\_click-link\_click-button\_click-checkboxes\_click-option\_click-dialog & 0.980 & 1.000 & 0.980 & -- & 0.140 & 0.360 & 0.100 & 0.300 & 0.480 & 0.000 & 0.000 & 0.000 & 0.000 & 0.000 & 0.000 & 0.000 & 0.000 \\
\midrule
\textbf{Average (n-way)} & \textbf{0.988} & \textbf{1.000} & \textbf{0.980} & -- & \textbf{0.130} & \textbf{0.158} & \textbf{0.234} & \textbf{0.222} & \textbf{0.584} & \textbf{0.000} & \textbf{0.074} & \textbf{0.074} & \textbf{0.000} & \textbf{0.244} & \textbf{0.244} & \textbf{0.014} & \textbf{0.218} \\
\midrule
click-checkboxes-transfer\_multi-layouts\_email-inbox-forward-nl-transition & 0.720 & 0.940 & 0.823 & -- & 0.000 & 0.000 & 0.000 & 0.000 & 0.000 & 0.000 & 0.000 & 0.000 & 0.000 & 0.000 & 0.000 & 0.220 & 0.440 \\
click-option\_login-user-transition & 1.000 & 1.000 & 1.000 & -- & 0.000 & 0.000 & 0.000 & 0.000 & 0.000 & 0.000 & 0.000 & 0.000 & 0.000 & 0.000 & 0.000 & 0.000 & 0.000 \\
click-option\_multi-layouts\_click-widget\_login-user-popup-transition & 0.480 & 0.940 & 0.823 & -- & 0.000 & 0.000 & 0.000 & 0.000 & 0.000 & 0.000 & 0.000 & 0.000 & 0.000 & 0.000 & 0.000 & 0.000 & 0.000 \\
login-user\_navigate-tree-transition & 0.860 & 0.960 & 0.820 & -- & 0.000 & 0.000 & 0.000 & 0.000 & 0.460 & 0.000 & 0.000 & 0.000 & 0.340 & 0.000 & 0.340 & 0.000 & 0.000 \\
login-user-popup\_email-inbox-forward-nl-turk-transition & 0.639 & 1.000 & 0.980 & -- & 0.000 & 0.000 & 0.200 & 0.000 & 0.120 & 0.000 & 0.000 & 0.000 & 0.000 & 0.000 & 0.000 & 0.700 & 0.560 \\
\midrule
\textbf{Average (transition)} & \textbf{0.740} & \textbf{0.968} & \textbf{0.889} & -- & \textbf{0.000} & \textbf{0.000} & \textbf{0.040} & \textbf{0.000} & \textbf{0.116} & \textbf{0.000} & \textbf{0.000} & \textbf{0.000} & \textbf{0.068} & \textbf{0.000} & \textbf{0.068} & \textbf{0.184} & \textbf{0.200} \\
\midrule
click-button\_click-tab-2-hard & 0.760 & 0.960 & 0.780 & -- & 0.080 & 0.120 & 0.260 & 0.340 & 0.580 & 0.280 & 0.040 & 0.280 & 0.960 & 0.320 & 0.960 & 0.260 & 0.280 \\
click-button-sequence\_use-autocomplete & 0.580 & 0.980 & 0.880 & -- & 0.200 & 0.060 & 0.640 & 0.160 & 0.820 & 0.000 & 0.000 & 0.000 & 0.420 & 0.000 & 0.420 & 0.000 & 0.000 \\
click-checkboxes-soft\_enter-password & 0.720 & 1.000 & 0.784 & -- & 0.000 & 0.000 & 0.000 & 0.000 & 0.820 & 0.000 & 0.000 & 0.000 & 0.000 & 0.000 & 0.000 & 0.000 & 0.000 \\
click-checkboxes-soft\_multi-layouts & 0.518 & 0.940 & 0.672 & -- & 0.000 & 0.000 & 0.000 & 0.000 & 0.120 & 0.000 & 0.300 & 0.000 & 0.000 & 0.820 & 0.820 & 0.000 & 0.000 \\
click-dialog\_search-engine & 1.000 & 1.000 & 1.000 & -- & 0.100 & 0.020 & 0.000 & 0.000 & 0.920 & 0.000 & 0.000 & 0.000 & 0.000 & 0.000 & 0.000 & 0.000 & 0.000 \\
click-dialog-2\_click-widget & 0.980 & 1.000 & 1.000 & -- & 0.000 & 0.000 & 0.000 & 0.040 & 0.020 & 0.100 & 0.000 & 0.100 & 0.180 & 0.000 & 0.180 & 0.160 & 0.100 \\
click-dialog-2\_login-user-popup & 0.680 & 1.000 & 0.980 & -- & 0.000 & 0.000 & 0.000 & 0.000 & 0.060 & 0.000 & 0.000 & 0.000 & 0.000 & 0.000 & 0.000 & 0.000 & 0.000 \\
click-widget\_click-checkboxes-soft & 0.706 & 1.000 & 0.800 & -- & 0.080 & 0.100 & 0.060 & 0.200 & 0.600 & 0.000 & 0.000 & 0.000 & 0.000 & 0.000 & 0.000 & 0.100 & 0.140 \\
enter-date\_login-user & 0.960 & 1.000 & 1.000 & -- & 0.000 & 0.020 & 0.000 & 0.060 & 0.500 & 0.000 & 0.000 & 0.000 & 0.000 & 0.000 & 0.000 & 0.380 & 0.380 \\
use-autocomplete\_click-dialog & 0.580 & 0.980 & 0.880 & -- & 0.000 & 0.080 & 0.000 & 0.080 & 0.060 & 0.000 & 0.000 & 0.000 & 0.000 & 0.000 & 0.000 & 0.000 & 0.000 \\
\midrule
\textbf{Average (two-way, easy-medium)} & \textbf{0.748} & \textbf{0.986} & \textbf{0.878} & -- & \textbf{0.046} & \textbf{0.040} & \textbf{0.096} & \textbf{0.088} & \textbf{0.450} & \textbf{0.038} & \textbf{0.034} & \textbf{0.038} & \textbf{0.156} & \textbf{0.114} & \textbf{0.238} & \textbf{0.090} & \textbf{0.090} \\
\textbf{Average (total)} & \textbf{0.891} & \textbf{0.990}& \textbf{0.941} & -- & \textbf{0.179} & \textbf{0.225} & \textbf{0.258} & \textbf{0.287} & \textbf{0.560} & \textbf{0.178} & \textbf{0.154} & \textbf{0.254} & \textbf{0.295} & \textbf{0.225} & \textbf{0.414} & \textbf{0.206} & \textbf{0.295} \\
\bottomrule
\end{tabular}
}
\end{center}
\caption{
Optimal success rate on \compwob{} (synthetically calculated as a product of base task success rate in MiniWoB).
}
\label{tab:per_task_compwob_results_optimal}
\end{table}
\end{landscape}
\end{tiny}

\clearpage
\section{Prompted Language Model Agents with Oracle Exemplars}
\label{appendix:oracle}

In this paper, we assume, as a realistic and important constraint, that: (1) \textbf{prompted or finetuned LMAs are deployed as a service in the real world}; (2) \textbf{users are only allowed to interact with the agents via providing task instruction}, and (3) \textbf{not allowed to intervene the backend system or prompts}. Providing exemplars or preparing finetuning demonstrations for every compositional problem is infeasible given the huge space of web automation problems. Moreover, it significantly hinders the user experience that the agents would be good at some specific format of instructions and that their violation of those affects the performance even if semantically the same.

\begin{figure*}[ht]
\centering
\includegraphics[width=\linewidth]{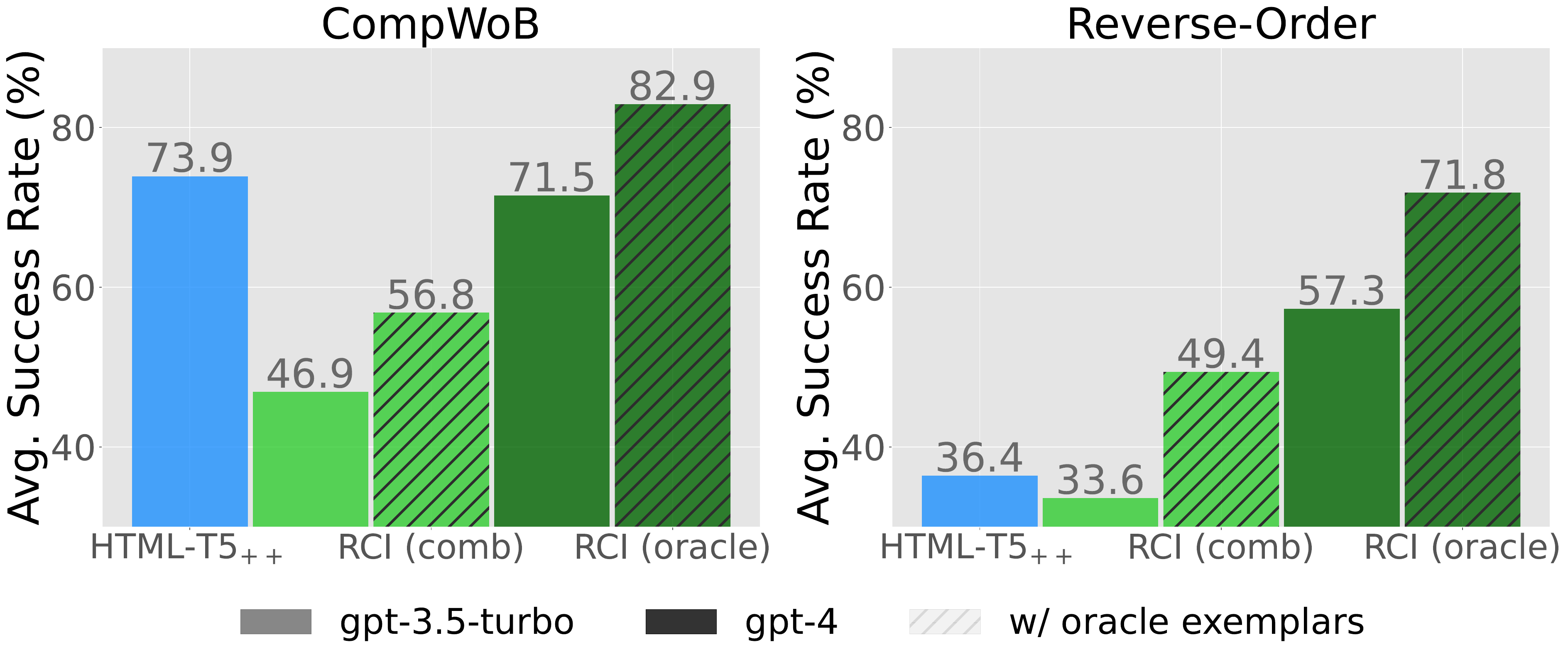}
\vskip -0.1in
\caption{Average success rate of LMAs in 20 two-way tasks from CompWoB.
RCI with \texttt{gpt-4} achieves the best performance when the oracle exemplars are provided (82.9\%) in the prompt.
While oracle demonstrations help improve performance, they could not fully resolve the issues from reverse-order instructions (for instance, 82.9\% $\rightarrow$ 71.8\% in RCI with \texttt{gpt-4}).
}
\label{fig:oracle_twoway}
\end{figure*}

As a sanity check of the environments and baseline agents, we here demonstrate that prompted LMAs can change their behaviors adaptively by modifying their prompts and demonstrations.
We evaluate the performance of RCI with \texttt{gpt-4} and oracle exemplars on 20 two-way tasks in CompWoB. We provide two demonstrations per task in the prompts.
\autoref{fig:oracle_twoway} shows that RCI with \texttt{gpt-4} and oracle exemplars achieves 82.9\% success rate, which is the best among the baselines, such as HTML-T5++ (73.9\%), RCI with \texttt{gpt-3.5-turbo} and oracle exemplars (56.8\%),  RCI with combination exemplars (\texttt{gpt-3.5-turbo}: 46.9\%, \texttt{gpt-4}: 71.5\%). See \autoref{tab:per_task_compwob_results} for other baselines. This ensures that the tasks are feasible, and that if a prompt includes how to perform on compositional tasks, the performance gets better.
In contrast, while oracle demonstrations help improve performance, they could not fully resolve the issues from reverse-order instructions.

Through the experiments in \cref{sec:results}, we have intended to shed light on their zero-shot generalization capabilities of dealing with unknown sequential-task compositions. For reliable and robust web automation agents, we might need to leverage both prompting and finetuning approaches at the different development stages. For instance, prompted LMAs might work well as adaptive automated data collectors to the novel domains and finetuned LMAs as deployed agents for service.

\clearpage
\section{Task Complexity Analysis with Language Model Agents}
\begin{figure*}[htb]
\centering
\includegraphics[width=\linewidth]{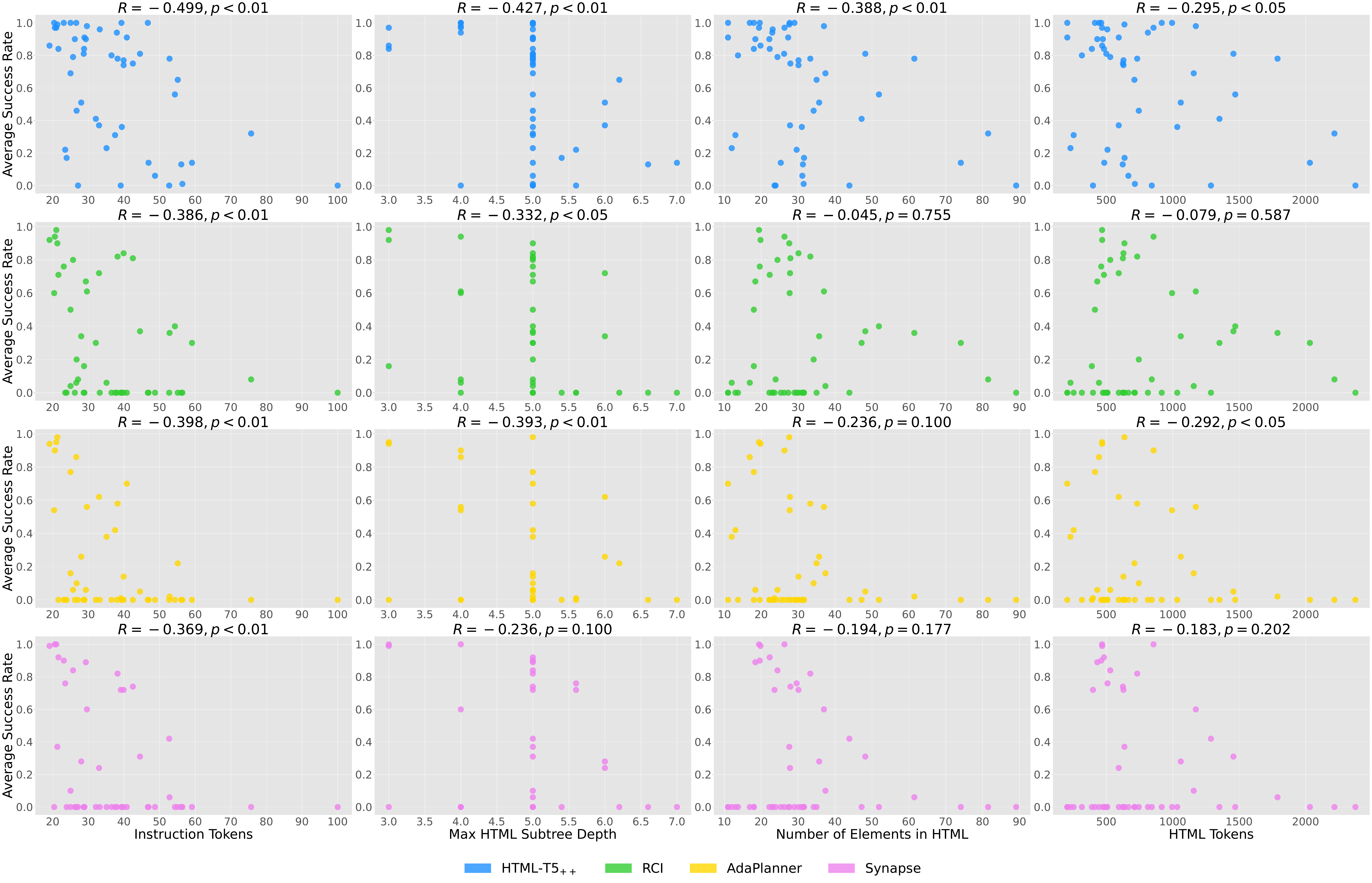}
\vskip -0.1in
\caption{
2D-scatter plots between the success rate for each LMA (y-axis) and each statistic of compositional task (x-axis), such as the number of instruction tokens, max depth of HTML subtrees, the number of elements in HTML, and the number of HTML tokens.
The results imply that while all the language model agents (HTML-T5++, RCI, AdaPlanner, Synapse) show negative correlations between the success rate and instruction tokens with statistical significance, the trends for other statistics may differ among the methods.
}
\label{fig:task_complexity_method}
\end{figure*}

Following the recent work in deep reinforcement learning literature~\citep{furuta2021pic}, we measure average performance as a proxy of oracle task solvability. If many kinds of agents perform poorly, such tasks are regarded as challenging. This can shed light on``task'' or ``environment'' themselves, rather than ``language model agents'' or ``prompting methods'', while such an analysis has been overlooked so far. Additionally, while the trend with average might be the same, distributional characteristics for each language model agent could be different. We extend our analysis by reporting the individual performances of each agent in \autoref{fig:task_complexity_method}. Our results still indicate that while all the language model agents (HTML-T5++, RCI, AdaPlanner, Synapse) often show negative correlations between the success rate and instruction tokens or max subtree depth with statistical significance, the trends for other statistics may differ among the methods.
\label{appendix:per_method_complexity}

\clearpage
\section{Finetuning May Generally Outperform Prompting in Compositional Tasks}
\label{appendix:lastletter}
\update{In this section, we examine the capability of finetuned LLMs and prompted LLMs for compositional generalization in natural language reasoning tasks, rather than agentic tasks.
We employ Last Letter Concatenation~\citep{wei2022cot,zhou2023leasttomost} as a benchmark, where if LLMs take \textit{``hans, structure''} as a question, LLMs are asked to answer it as \textit{The last letter of ``hans'' is ``s''. The last letter of ``structure'' is ``e''. Considering ``s'', ``e'' leads to ``se''. So, ``hans, structure'' outputs ``se''}.
This reasoning problem exhibits a compositional nature as we increase the number of words provided as a question.
For finetuned LLMs, we adopt 11B-parameter Flan-T5-XXL~\citep{chung2022flant5,2020t5} as a base model, and prepare 2000 examples per $n$-letter for a training dataset.
For prompted LLMs, we employ Flan-PaLM-8B, 62B, and 540B~\citep{chung2022flant5,Chowdhery2022palm}, and prepare randomized 4-shot CoT prompts from the same training dataset. We randomly use one of them during inference.
Furthermore, we use up to 8 letters as a training dataset and few-shot exemplars and also use 9-12 letters as a hold-out test dataset to investigate the compositional generalization in natural language reasoning tasks.}

\update{\autoref{tab:last_letter_concatenation} shows that finetuned LLM (Flan-T5-XXL) almost always outperforms prompted LLM (Flan-U-PaLM-540B) on average (76\% v.s. 44\% in an exact match) and also in both in-distribution (2-8 letters) and out-of-distribution (9-12 letters) settings.
This implies that finetuned LLMs may generally outperform prompted LLMs in compositional tasks, which can also lead to the better performance of finetuned LMAs than prompted LMAs in sequential web automation tasks, as shown in the main text.}

\begin{table}[hb]
\begin{center}
\begin{small}
\scalebox{0.92}{
\begin{tabular}{lrrrrrrr|rrrr|r}
\toprule
 & \multicolumn{12}{c}{\textbf{\# of Letters}} \\
\cmidrule(r){2-13}
\textbf{Models} & 2 & 3 & 4 & 5 & 6 & 7 & 8 & 9 & 10 & 11 & 12 & Ave. \\
\midrule
\texttt{Flan-PaLM-8B} (Prompted) & 0.10 & 0.01 & 0.00 & 0.00 & 0.00 & 0.00 & 0.00 & 0.00 & 0.00 & 0.00 & 0.00 & 0.01 \\
\texttt{Flan-PaLM-62B} (Prompted) & 0.48 & 0.16 & 0.05 & 0.03 & 0.00 & 0.00 & 0.00 & 0.00 & 0.00 & 0.00 & 0.00 & 0.07 \\
\texttt{Flan-U-PaLM-540B} (Prompted) & \textbf{1.00} & \textbf{1.00} & 0.75 & 0.76 & 0.57 & 0.41 & 0.16 & 0.08 & 0.06 & 0.00 & 0.01 & 0.44 \\
\midrule
\texttt{Flan-T5-XXL} (11B; Finetuned) & \textbf{1.00} & 0.98 & \textbf{0.99} & \textbf{0.98} & \textbf{0.98} & \textbf{0.96} & \textbf{0.96} & \textbf{0.83} & \textbf{0.47} & \textbf{0.20} & \textbf{0.05} & \textbf{0.76} \\
\bottomrule
\end{tabular}
}
\end{small}
\end{center}
\vskip -0.15in
\caption{
    \update{The performance of finetuning (Flan-T5-XXL~\citep{chung2022flant5,2020t5}, 11B parameters) and prompting (Flan-PaLM-8B/62B/540B~\citep{chung2022flant5,Chowdhery2022palm}) in Last Letter Concatenation~\citep{wei2022cot,zhou2023leasttomost}. We measure the performance with an exact match.
    We use up to 8 letters as a training dataset and few-shot exemplars and also use 9-12 letters as a hold-out test dataset to investigate the compositional generalization in natural language reasoning tasks.
    The results show that finetuned LLM (Flan-T5-XXL) almost always outperforms prompted LLM (Flan-U-PaLM-540B) in both in-distribution (2-8 letters) and out-of-distribution (9-12 letters) settings.}
}
\label{tab:last_letter_concatenation}
\end{table}

\end{document}